\documentclass[lettersize,journal]{IEEEtran}
\usepackage{amsmath,amsfonts}
\usepackage{array}
\usepackage[caption=false,font=normalsize,labelfont=sf,textfont=sf]{subfig}
\usepackage{textcomp}
\usepackage{stfloats}
\usepackage{url}
\usepackage{verbatim}
\usepackage{graphicx}
\usepackage{balance}
\usepackage{xcolor}
\usepackage{amsmath}
\usepackage{hyperref}
\usepackage{bm}
\usepackage{booktabs}
\usepackage{enumitem}
\usepackage{algorithm}
\usepackage{algorithmic}
\usepackage{footnote}
\usepackage{multirow}
\usepackage{cleveref}
\usepackage{etoolbox}
\usepackage{amsthm}
\usepackage{subcaption}
\usepackage{flushend} 
\usepackage{xspace}

\newtheorem{mydef}{Definition}

\def\BibTeX{{\rm B\kern-.05em{\sc i\kern-.025em b}\kern-.08em
    T\kern-.1667em\lower.7ex\hbox{E}\kern-.125emX}}

\begin{document}
\author{
BinXu Wu\textsuperscript{2,3}, TengFei Zhang\textsuperscript{2}, Chen Yang\textsuperscript{2,3}, JiaHao Wen\textsuperscript{2,3}, HaoCheng Li\textsuperscript{2}, JingTian Ma\textsuperscript{3}, Zhen Chen*\textsuperscript{1,2}, JingYuan Wang*\textsuperscript{3} 
\thanks{\textsuperscript{1}Tsinghua University, Beijing, China} 
\thanks{\textsuperscript{2}Encosmart, Beijing, China} 
\thanks{\textsuperscript{3}Beihang University, Beijing, China} 
}

\newcommand{\inputsection}[1]{\input{sections_en/#1}}
\newcommand{\name}{SAGE}

\newcommand{\ie}{i.e.\xspace}
\newcommand{\eg}{e.g.\xspace}

\title{SAGE: State-Aware Guided End-to-End Policy for Multi-Stage Sequential Tasks via Hidden Markov Decision Process}

\markboth{}%
{作者姓氏: 简短标题}

\maketitle

\begin{abstract}
Multi-stage sequential (MSS) robotic manipulation tasks are prevalent and crucial in robotics. They often involve state ambiguity, where visually similar observations correspond to different actions. We present SAGE, a state-aware guided imitation learning framework that models tasks as a Hidden Markov Decision Process (HMDP) to explicitly capture latent task stages and resolve ambiguity. We instantiate the HMDP with a state transition network that infers hidden states, and a state-aware action policy that conditions on both observations and hidden states to produce actions, thereby enabling disambiguation across task stages. To reduce manual annotation effort, we propose a semi-automatic labeling pipeline combining active learning and soft label interpolation. In real-world experiments across multiple complex MSS tasks with state ambiguity, SAGE achieved 100\% task success under the standard evaluation protocol, markedly surpassing the baselines. Ablation studies further show that such performance can be maintained with manual labeling for only about 13\% of the states, indicating its strong effectiveness.
\end{abstract}


\section{Introduction}
\IEEEPARstart{R}{obotic} manipulation tasks have attracted significant attention due to their broad  applications. Vision-based strategies have been widely adopted~\cite{savva2019habitat}, and have demonstrated remarkable performance across a variety of real-world scenarios~\cite{chi2023diffusion,Self-Supervised-Visuomotor,DP3-Visuomotor,CAPR-Visuomotor,ze2023gnfactor}. However, a particular class of tasks—\emph{Multi-Stage Sequential} (MSS) tasks—introduces distinctive challenges to vision-based policies. MSS tasks are characterized by a sequence of interdependent stages that must be executed in a prescribed temporal order, often requiring the policy to perform long-horizon reasoning, retain contextual information from prior steps, and ensure coherent progression across successive stages.

\begin{figure}[t]
\centering
\includegraphics[width=0.5\textwidth]{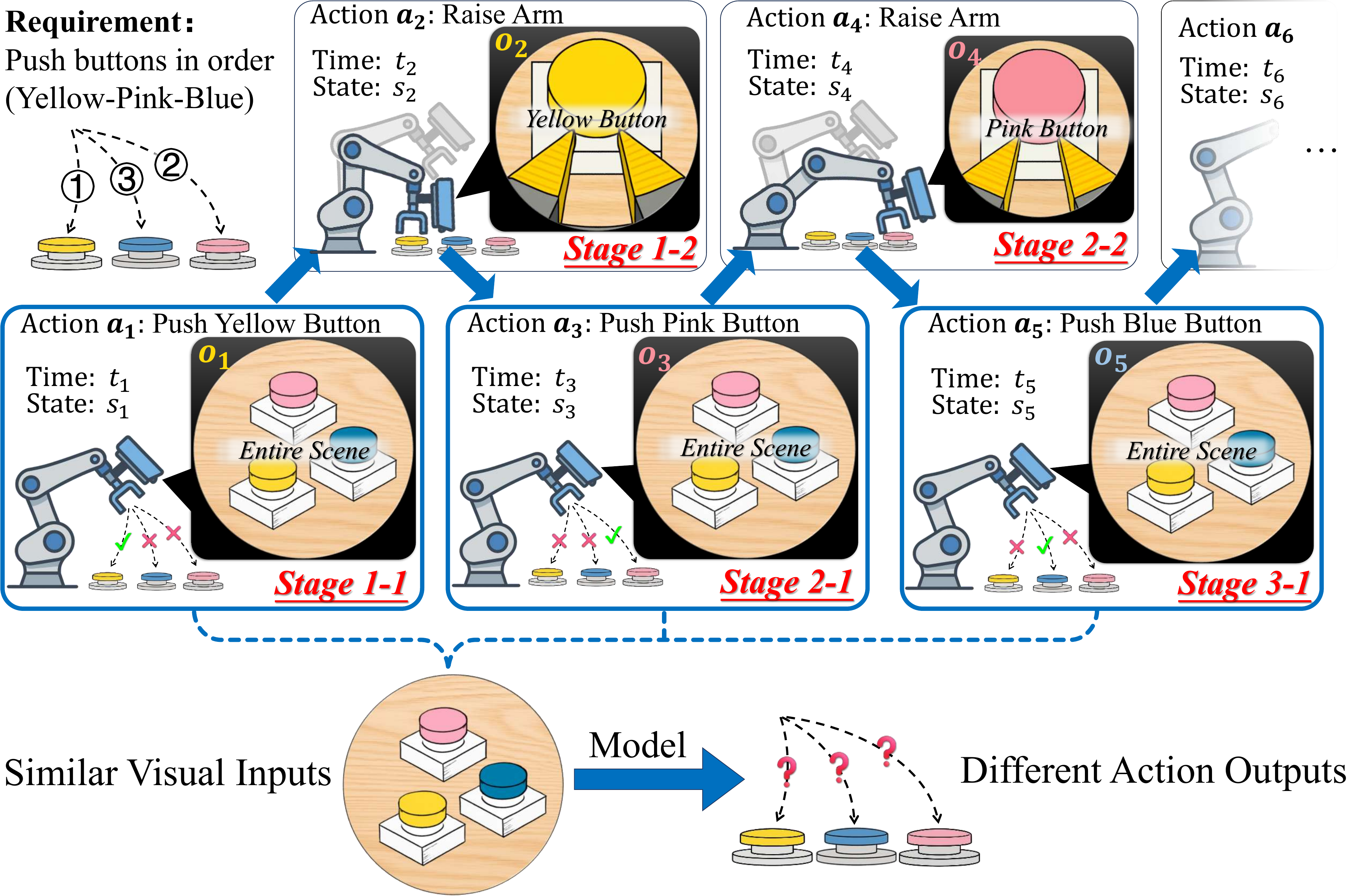}
\vspace{-0.5cm}
\caption{\textbf{Push Buttons} task, where the robot must press buttons in order (Yellow–Pink–Blue). Since buttons look almost the same before and after being pushed, it is hard to tell whether a button has already been pushed.}

\label{fig:introduction}
\vspace{-0.5cm}
\end{figure}

In many MSS tasks, conventional vision-based policies struggle in scenarios involving \textit{state ambiguity}. In such cases, visually similar observations may correspond to different actions, resulting in ambiguity during action selection. An illustrative case is the \textit{Push Buttons} task shown in Fig.~\ref{fig:introduction}. The visual observations at stages 1-1, 2-1, and 3-1 are nearly indistinguishable; however, the correct action—pressing the yellow, pink, or blue button—requires knowledge of the current task stage to be correctly determined. This requires the policy to map similar observations to distinct actions, a phenomenon we refer to as \textit{State Ambiguity}. Similar challenges also arise in other real-world contexts, such as assessing whether a container has been filled in a warehouse packaging task, or judging whether a cloth is wet or dry during household cleaning. These examples highlight the inherent difficulty of resolving state ambiguity when relying solely on visual input.

To handle state ambiguity in MSS tasks, existing methods mainly fall into two categories: Memory-based approaches and hierarchical task decomposition. (1) In robotic manipulation, memory-based methods use models like recurrent neural networks~\cite{mandlekar2022matters}, attention mechanisms~\cite{fang2025sam2act}, Transformer-XL~\cite{tianci2024transformer} to capture historical context, as incorporating earlier observations can help distinguish visually similar states that lead to ambiguity. Although these methods are flexible, they often struggle with redundant information, high computational cost, and difficulties in deciding how much history to retain. (2)Hierarchical approaches, on the other hand, structure the policy into multiple levels of controllers. A high-level controller manages stage transitions, while low-level policies are responsible for executing specific actions~\cite{luo2024multistage,marzinotto2014towards,silver2023learning,wu2024afforddp}. While this structure helps reduce ambiguity, designing the high-level controller typically requires extensive manual effort. For example, some methods use a nine-layer decision tree to handle transitions~\cite{marzinotto2014towards}. Moreover, since low-level modules directly execute action primitives, they are often constrained to be simple and modular, which limits flexibility. Transitions between primitives further introduce delays, reducing execution efficiency and stability in real-world scenarios.

To address the issue of \textit{State Ambiguity}, we propose \textbf{SAGE}, a \textbf{S}tate-\textbf{A}ware \textbf{G}uided \textbf{E}nd-to-End Imitation Learning framework based on the Hidden Markov Decision Process (HMDP). Specifically, in Section~\ref{sec:Theoretical}, we provide a theoretical analysis. The key idea is to treat observations as partial manifestations of a latent environment state and to explicitly model this state as a hidden variable. With this formulation, the HMDP can distinguish between different underlying physical states that share similar visual appearances, thereby resolving ambiguity. The HMDP consists of two components: \emph{hidden state estimation} and \emph{decision-making agent}. In Section~\ref{sec:HMDP-Real-world}, guided by the theoretical formulation, we implement these components as two neural networks. The \emph{state transition network} infers hidden states from the current observation together with the previously estimated state, while the \emph{state-aware action policy} generates actions by conditioning jointly on visual observations and the inferred states. We integrate them into an End-to-End architecture and train the framework using actions from expert demonstrations and human-labeled states as supervision. Furthermore, as detailed in Section~\ref{sec:auto_tag}, to reduce the manual annotation cost of state labels required by supervised learning, we propose a semi-automatic labeling pipeline that integrates \textit{active learning} with \textit{soft label interpolation}. It substantially reduces manual labeling effort by annotating only a small subset of representative segments and automatically labeling the remaining data.

Extensive real-world experiments on three MSS tasks, all of which involve state ambiguity, were conducted to evaluate the effectiveness of SAGE. The results show that it achieves up to a \textbf{100\%} stage success rate under standard evaluation settings, significantly outperforming competitive baselines. Our method remains robust under visually distracting conditions and continuous execution, completing 50-step sequences without error. Ablation studies show that our semi-automatic annotation strategy in SAGE achieves full success with only 13\% of episodes manually labeled, demonstrating efficient annotation. These results validate the effectiveness and generality of our approach in tackling real-world state ambiguity. 

Therefore, we propose a unified imitation learning framework. Our main contributions are as follows:
\begin{itemize}
    \item To the best of our knowledge, we are the first to formulate MSS tasks with \textit{state ambiguity} as a Hidden Markov Decision Process, offering a principled framework to address this challenge.
    \item We realize the HMDP formulation through two dedicated neural modules for hidden state inference and state-aware action generation, which are jointly integrated into an End-to-End training pipeline.
    \item We propose a semi-automatic state annotation strategy that substantially reduces the human effort required to provide state supervision signals for the HMDP.
    \item We conduct extensive real-world experiments on multiple MSS tasks with state ambiguity, demonstrating the superior performance and robustness of SAGE.
\end{itemize}

\section{Related Work}
\subsection{Visual Imitation Learning}

Visual imitation learning has been widely applied in robotic manipulation tasks~\cite{shridhar2022peract, lu2024manigaussian, goyal2024rvt2, zhu2023viola}. In recent years, large language models have shown potential for improving policy generalization~\cite{ahn2022saycan,liang2023code,huang2023inner, huang2022language, liang2022code}; however, such approaches often incur significant computational overhead and high deployment costs, making them difficult to apply directly in real-world systems. Other works explore alternative directions such as transferring human hand-object motion from videos to robot hands~\cite{qin2022dexmv}, enabling one-shot imitation by decomposing tasks into coarse and fine phases~\cite{johns2021coarse}, or introducing skill prompts for hierarchical continual learning~\cite{DBLP:journals/corr/abs-2504-15561,ma2024hierarchical}.
However, these approaches still struggle to generate diverse and scalable low-level actions, which motivates the development of diffusion-based visuomotor policies.

Diffusion-based visuomotor policies, exemplified by Diffusion Policy~\cite{chi2023diffusion}, generate actions via iterative denoising and have shown strong performance on short-horizon tasks. Recent extensions enhance their applicability to more complex scenarios: DP3~\cite{DP3-Visuomotor} incorporates 3D point clouds for better depth reasoning, AnchorDP3~\cite{3DAffordanceDP} anchors keyposes to affordances for dual-arm manipulation, HDP~\cite{DBLP:journals/trob/WangLCX25} leverages contact-guided trajectory planning via hierarchy, and RDP~\cite{DBLP:journals/corr/abs-2503-02881} introduces visual-tactile feedback for reactive control. Despite these advances, most approaches remain reactive and lack explicit modeling of temporal task structure, making them susceptible to state ambiguity in long-horizon settings.

\subsection{Memory and Hierarchical Methods}

\label{subsec:related_work}
To address state ambiguity in MSS tasks, effective long-horizon modeling remains a central challenge in robotic manipulation. Existing approaches can be broadly categorized into two classes: Memory-based methods and hierarchical task decomposition. The former captures historical context using sequential or spatial encodings, such as recurrent neural networks~\cite{mandlekar2022matters}, attention mechanisms~\cite{fang2025sam2act}, Transformer-XL~\cite{tianci2024transformer}, and state-space models like Mamba~\cite{gu2023mamba}. Some methods also incorporate spatial memory using Gaussian splatting to encode semantic and geometric structure in manipulation tasks~\cite{GaussianMemory,GaussianMutliStage,guhur2023instruction}. While offering flexibility, these methods often suffer from excessive and redundant information, high computational overhead, and difficulty in determining an appropriate history length. RDMemory~\cite{StillInMind} introduces explicit object-centric memory to handle occlusions, but remains limited in capturing scene-level progression and task-stage dynamics crucial for resolving state ambiguity.

Another line of work adopts hierarchical architectures that decompose complex tasks into multiple sub-stages \cite{silver2023learning, HVIL, gupta2019relay}. A high-level controller is responsible for stage recognition and switching, while low-level policies execute specific actions. For example, the Multi-Stage Cable Routing method~\cite{luo2024multistage} divides the manipulation process into sub-tasks such as Route, Pickup, and Perturb, each handled by separate models. Behavior tree methods manage stages through manually designed multi-layer decision logic~\cite{marzinotto2014towards, Behavior-Tree}. Related work also addresses long-horizon robotic tasks by structuring planning into symbolic abstractions~\cite{liang2022sem,LearningSymbolicTAMP,kim2020relational}, action sequence prediction~\cite{driess2020deep}, or belief-space reasoning~\cite{garrett2020online}. However, hierarchical approaches typically require task-specific high-level controllers, limiting their generality. Low-level primitives are constrained to remain overly simple and modular, which limits flexibility. In addition, switching between primitives can introduce execution delays, reducing efficiency in practice.

\subsection{Efficient Data Labeling in Robot Learning}

In robotic learning, high-quality data is often expensive and time-consuming to collect. Recent works have explored efficient ways to acquire large-scale robot trajectories, including self-supervised data collection with visuo-tactile feedback~\cite{fu2023safe}, self-improving agents~\cite{self-improving}, scalable motion capture systems~\cite{DexCap}, synthetic demonstration generation~\cite{DemoGen,zhou2024robodreamer,mandlekar2023mimicgen}, and influence-guided data curation~\cite{CUPID}. While these methods provide diverse manipulation trajectories, they are not tailored for tasks that require additional structured labels, such as hidden states in our work.

To further reduce the annotation burden, two main directions have been explored.
The first is prioritizing informative samples, typically through active learning strategies. Representative approaches include adversarial selection~\cite{sinha2019variational} and representativeness-based sampling for better feature coverage~\cite{sener2018active}. Beyond these classical methods, techniques like SCIZOR~\cite{SCIZOR} further improve dataset quality by filtering suboptimal or redundant samples, thereby enhancing diversity and coverage even outside standard active learning paradigms. The second direction focuses on enhancing supervision quality. For instance, label smoothing methods~\cite{rethink} have been shown to improve generalization under noisy supervision. In addition, recent works explore automatic labeling and annotation augmentation: NILS~\cite{NILS} leverages foundation models for zero-shot labeling of unstructured data, while RoCoDA~\cite{RoCoDA} augments demonstrations with counterfactual examples to improve policy generalization.

Building on these insights, we propose a hybrid labeling strategy that combines active learning sampling with soft label interpolation to enable efficient annotation.

\section{Theoretical Analysis}
\label{sec:Theoretical}
\subsection{Problem Definition}

The objective of this study is to train a control model via \emph{End-to-End imitation learning} for \emph{vision-based robotic manipulation} in \emph{multi-stage sequential tasks}. This subsection briefly outlines the core components of our framework.

\textbf{Vision-based Robotic Manipulation:} As illustrated in Fig.~\ref{fig:introduction}, vision-based robotic manipulation typically involves a robotic arm equipped with one or more cameras for external perception. The control model processes visual observations as input and generates control commands to execute the required tasks.

\textbf{Multi-stage Sequential (MSS) Tasks:} MSS tasks consist of temporally ordered sub-tasks or actions. They require long-horizon planning, memory of past states, and smooth transitions across stages. Typical examples include object rearrangement, tool use, and pick-navigate-place operations. MSS tasks are common in household robotics, industrial automation, and warehouse logistics.

\textbf{End-to-end Imitation Learning:} End-to-end imitation learning trains a deep learning model to map raw sensory inputs (e.g., images, proprioception) directly to low-level control actions by imitating expert demonstrations, without relying on intermediate representations or modular pipelines. Training data consist of images from the robot’s camera paired with expert-provided actions. The goal is to reproduce expert-like behaviors. This approach leverages deep learning’s capacity to model complex perception-action mappings and is well-suited for tasks where hand-crafted policies are difficult to design.

\subsection{Markov Decision Process and State Ambiguity}

The control model for a vision-based robotic manipulation is typically formulated as a Markov Decision Process (MDP).

\begin{mydef}[Markov Decision Process]~\label{def:mdp}
    A Markov Decision Process is a stochastic process with three variables:
    \begin{itemize}
    \item{\em State:} At time $t$, the environment is in state $s_t$, with $s_0$ as the initial state.
    
    \item{\em Action:} Given $s_t$, the agent takes action $a_t$.
    
    \item{\em Reward:} After action $a_t$, the agent receives reward $r_t$.
\end{itemize}
    These variables are governed by the following functions:
    \begin{itemize}
    \item{\em Agent:} The agent follows a policy $\pi$ that maps state $s_t$ to action $a_t$, i.e., $a_t = \pi(s_t; \theta_\pi)$, where $\theta_\pi$ are parameters.
    
    \item{\em State Transition:} Describes the probability of transitioning from $s_t$ to $s_{t+1}$ under action $a_t$, i.e., $\mathrm{Pr}(s_{t+1} \mid s_t, a_t)$.
    
    \item{\em Reward Function:} Assigns reward $r_t$ based on state $s_t$ and action $a_t$, i.e., $r_t = R(s_t, a_t)$.
\end{itemize}

    The goal is to find $\theta_\pi$ that maximizes the total reward over the trajectory: { $\theta_\pi^{\ast} = \mathop{\arg\max}\limits_{\theta_\pi} \sum_{t=1}^{T} r_t$}.
\end{mydef}

In vision-based robotic manipulation, images from the camera and proprioceptive signals together constitute the environment state, and action $a_t$ represents the control signal to the arm. The agent acts as the controller, generating actions based on observation input. Under the End-to-End imitation learning framework, the agent parameter $\theta_\pi$ is trained via a reward function. A common reward is defined as the negative error between the agent’s and the expert’s actions:
\begin{equation*}\label{11}
  r_t(\theta_\pi) = - \Big\| \pi(s_t; \theta_\pi) - a'_t \Big\|,
\end{equation*}
where $a'_t$ is the expert action and action $a_t = \pi(s_t; \theta_\pi)$.


\textbf{State Ambiguity Problem in MDP.} Training a control agent with a standard MDP for multi-stage sequential tasks can lead to {\em state ambiguity} problem. To illustrate this problem, we present a toy example (see Fig.~\ref{fig:introduction}). Consider a task requiring the robot to push three buttons in the fixed order: Yellow, pink, blue. This task involves three stages:
\begin{itemize}
  \item {\em Stage 1}: Push the yellow button;
  \item {\em Stage 2}: Push the pink button;
  \item {\em Stage 3}: Push the blue button.
\end{itemize}
In each stage, the robot performs two actions: $i$) Raise the arm to view the entire scene, where the image includes all three buttons; $ii$) Lower the arm to push a button, where the image shows only one button. The corresponding MDP is described in the following table:
\begin{table}[h]
  \centering
  \begin{tabular}{c|c|c}
  \toprule
    & {State} & {Action} \\ \midrule
  $t_1$ & $s_1$ = $\mathrm{Entire\;Scene}$ & $a_1$ = $\mathrm{Push\;Yellow\;Button\; (Lower\;Arm)}$ \\
  $t_2$ & $s_2$ = $\mathrm{Yellow\;Button}$ & $a_2$ = $\mathrm{Raise\;Arm}$ \\
  $t_3$ & $s_3$ = $\mathrm{Entire\;Scene}$ & $a_3$ = $\mathrm{Push\;Pink\;Button\; (Lower\;Arm)}$ \\
  $t_4$ & $s_4$ = $\mathrm{Pink\;Button}$ & $a_4$ = $\mathrm{Raise\;Arm}$ \\
  $t_5$ & $s_5$ = $\mathrm{Entire\;Scene}$ & $a_5$ = $\mathrm{Push\;Blue\;Button\; (Lower\;Arm)}$ \\
  $t_6$ & $s_6$ = $\mathrm{Blue\;Button}$ & $a_6$ = $\mathrm{Raise\;Arm}$ \\
  \bottomrule
  \end{tabular}
\end{table}

In this setup, the agent faces a state ambiguity problem: At time steps $t_1$, $t_3$, and $t_5$, it receives the same observation $s_1 = s_3 = s_5$ \footnote{Strictly speaking, $s_t$ also includes proprioceptive signals such as joint configurations. We ignore them here because in this example the proprioceptive states at $s_1$, $s_3$, and $s_5$ are identical, and visual inputs typically dominate inference in vision imitation learning. Hence, the ambiguity can be illustrated using visual observations alone.} but must produce different actions $a_1 \neq a_3 \neq a_5$, \ie,
\begin{equation}
\begin{aligned}
    \pi(s_1) \neq \pi(s_3) \neq \pi(s_5),\quad \\
    s_1 = s_3 = s_5 = \mathrm{Entire\;Scene}.
\end{aligned}
\end{equation}
We define this as the \emph{State Ambiguity} problem:

\begin{mydef}[State Ambiguity]~\label{def:DA}
    Given an MDP $\{S, A, \pi, R\}$, if there exist two decision steps $t_1$ and $t_2$ such that $s_{t_1} = s_{t_2}$ but $\pi(s_{t_1}) \neq \pi(s_{t_2})$, and this discrepancy is necessary to maximize $R$, the MDP exhibits \emph{Decision Ambiguity}.
\end{mydef}

State ambiguity commonly arises in multi-stage tasks—such as assembling parts in order, performing staged surgical procedures, or pushing buttons in a fixed sequence—where relying solely on the current state $s_t$ fails to capture the necessary task context for correct action selection.

Some existing methods address this by extending the input to include a history of past states, \ie, $a_t = \pi(s_t, \ldots, s_{t-\tau})$, effectively turning the MDP into a higher-order Markov process~\cite{high-order-markov}. However, this approach has two major limitations: First, choosing the proper history length is difficult, as different task stages may require varying levels of historical context. An ideal agent would adapt its temporal scope dynamically—something hard to implement. Second, in vision-based control, states are images; handling long histories means processing many frames (\ie, long video segments), which incurs heavy computational costs and hinders practical deployment.

\subsection{Hidden Markov Decision Process} 
\label{subsec:HMDP}
\subsubsection{Formulation of HMDP}
\label{subsubsec:formulation}
To address the \emph{state ambiguity} problem in Def.~\ref{def:DA}, we extend the standard MDP framework to a \emph{Hidden Markov Decision Process} (HMDP).

\begin{mydef}[Hidden Markov Decision Process]~\label{def:HMDP}
    A hidden Markov decision process is a discrete-time stochastic model defined by four key variables:
    \begin{itemize}
        \item \emph{State}: At time $t$, the environment is in state $s_t$, with $s_0$ as the initial state. Unlike standard MDPs, $s_t$ is unobservable.
        \item \emph{Observation}: The agent receives a partial observation $o_t$ of the hidden state $s_t$.
        \item \emph{Action}: The agent takes action $a_t$ to influence the environment.
        \item \emph{Reward}: After action $a_t$, the agent receives reward $r_t$.
    \end{itemize}

    These variables interact through the following functions:
    \begin{itemize}
        \item \emph{State Inference}: The hidden state is estimated via
        \begin{equation}~\label{eq:state_inferring}
            \hat{s}_t = f(o_t, \hat{s}_{t-1}; \theta_f),
        \end{equation}
        where $\theta_f$ are the model parameters, and $\hat{s}_t$ denotes the model-predicted (estimated) state at time $t$.
        
        \item \emph{Agent}: The agent's policy $\pi$ maps the observation and inferred state to action:
        \begin{equation*}
            a_t = \pi(o_t, \hat{s}_t; \theta_\pi),
        \end{equation*}
        where $\theta_\pi$ are the policy parameters.
        
        \item \emph{Reward Function}: The reward is $r_t = R(s_t, o_t, a_t)$.
        
        \item \emph{State Transition}: The environment transitions from $s_t$ to $s_{t+1}$ according to $\mathrm{Pr}(s_{t+1} \mid s_t, a_t)$.
        
        \item \emph{Emission Probability}: Observations obey $\mathrm{Pr}(o_t \mid s_t)$.
    \end{itemize}

    The objective is to optimize $\theta_\pi$ to maximize reward:
    \begin{equation*}
        \theta_\pi^{\ast} = \mathop{\arg\max}\limits_{\theta_\pi} \sum_{t=1}^{T} r_t.
    \end{equation*}
\end{mydef}

The HMDP incorporates features of the Hidden Markov Model~\cite{hmm} into the MDP framework, where the true environment state is unobservable, and $o_t$ represents a partial observation of the underlying state $s_t$. The agent must infer an estimated state $\hat{s}_t$ from these observations.

\begin{figure}
    \centering
    \includegraphics[width=1\linewidth]{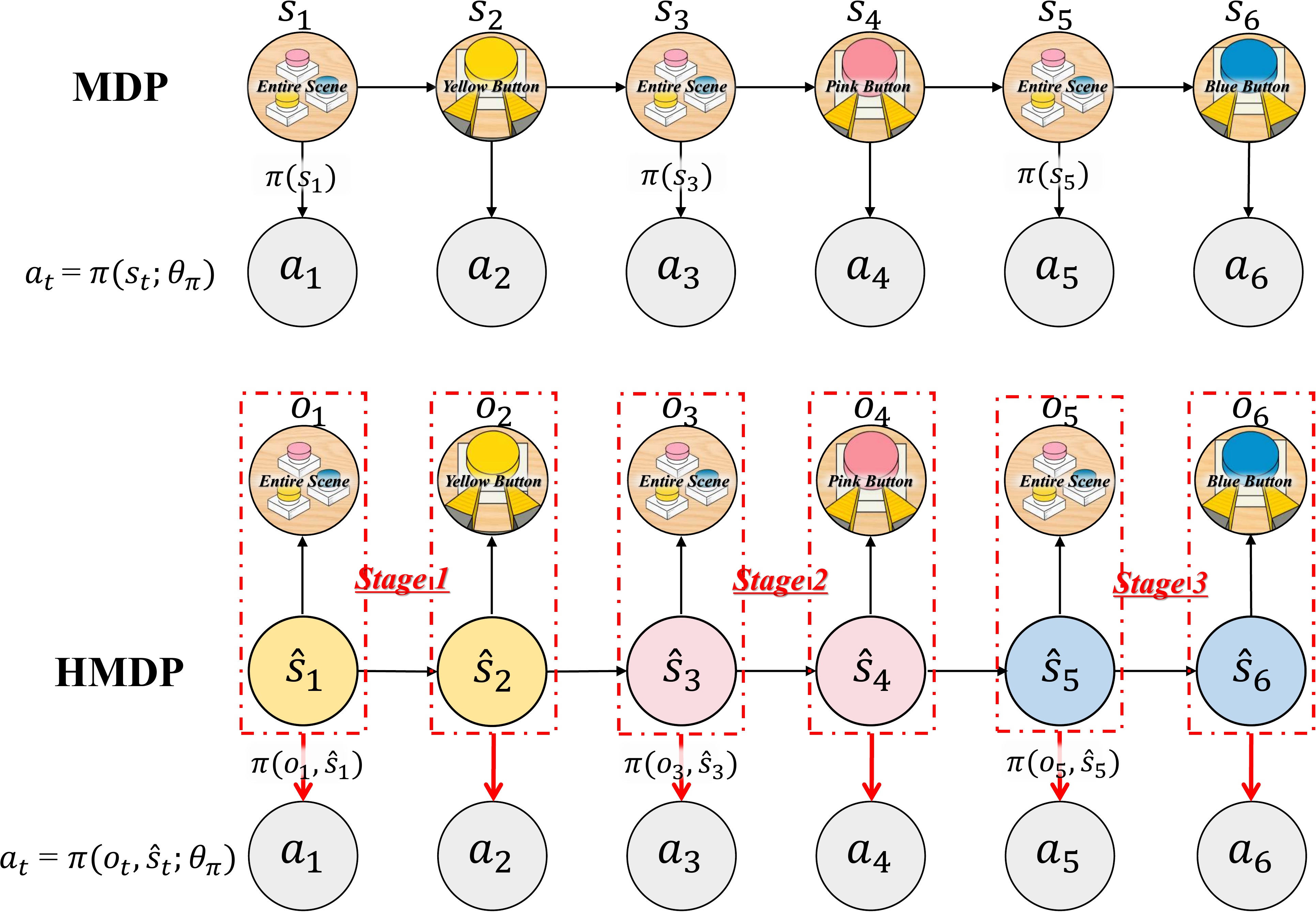}
    \vspace{-0.5cm}
    \caption{Comparison between Markov Decision Process (MDP) and Hidden Markov Decision Process (HMDP) in vision-based robotic manipulation control.}
    \vspace{-0.5cm}
    \label{fig:markov}
\end{figure}

\textbf{Eliminating State Ambiguity via HMDP.} Unlike standard MDPs, where $(\text{vision, proprioception})$ are treated as the full environment state $s_t$, HMDP regards them as partial observations $o_t$ of a hidden task stage, as shown in Fig.~\ref{fig:markov}. HMDP instead models the true state as the logical stage of a multi-stage task. This resolves state ambiguity: Even if $o_{t_1} = o_{t_2}$, the agent can produce different actions based on the inferred hidden states, i.e., $\pi(o_{t_1}, \hat{s}_{t_1}) \neq \pi(o_{t_2}, \hat{s}_{t_2})$ if $\hat{s}_{t_1} \neq \hat{s}_{t_2}$.

Using the “Push Buttons” task as an example (Fig.~\ref{fig:introduction} and Fig.~\ref{fig:markov}), if we define the hidden states as high-level task stages (e.g., ``Yellow Button Stage'', ``Pink Button Stage'', ``Blue Button Stage'') and treat camera images as observations, then the ambiguity in stages 1, 3, and 5 can be resolved as:
\begin{equation}
    \begin{aligned}
        \pi(& o_1, s_1) \neq \pi(o_3, s_3) \neq \pi(o_5, s_5), \\
        o_1 &= o_3 = o_5 = \mathrm{``Entire\;Scene"}, \\
        s_1 &= \mathrm{``Yellow\;Button\;Stage"}, \\
        s_3 &= \mathrm{``Pink\;Button\;Stage"}, \\
        s_5 &= \mathrm{``Blue\;Button\;Stage"}.
    \end{aligned}
\end{equation}
Thus, modeling the control system as an HMDP effectively eliminates state ambiguity in vision-based robotic tasks.

\subsubsection{Hidden State Estimation}
\label{subsubsec:infer}
In Eq.~\eqref{eq:state_inferring}, HMDP infers the hidden state sequence $(s_1, \ldots, s_t)$ from the observation sequence $(o_1, \ldots, o_t)$, corresponding to the {\em decoding problem} in Hidden Markov Models~\cite{hmm}. The objective is to maximize the conditional probability, where {$a_t = \pi(o_t, \hat{s}_t; \theta_\pi)$} :
\begin{equation}\label{eq:decoding}
    \begin{aligned}
        \!(\!\hat{s}_1, \!\ldots,\! \hat{s}_t\!)\!\! &= \!\!\mathop{\arg\max}\limits_{\{{s}_i\}_{i=1}^t} \mathrm{Pr}(s_1, \ldots, s_t \mid o_1, \ldots, o_t; \pi) \\
        &=\!\! \mathop{\arg\!\max}\limits_{\{{s}_i\}_{i=1}^t} \mathrm{Pr}(s_0)  \!\prod_{i=1}^{t} \!\mathrm{Pr}(o_i \!\mid \!s_i)\mathrm{Pr}(s_i\! \mid\! a_{i-1}, s_{i-1})
    \end{aligned}
\end{equation}

We adopt an approximation method to infer each hidden state $s_i$ in Eq.~\eqref{eq:decoding}. Assuming the previous estimate is $\hat{s}_{i-1}$ and the corresponding action is $a_{i-1} = \pi(o_{i-1}, \hat{s}_{i-1})$, the $i$-th term becomes:
\begin{equation}\label{eq:term_probability}
  \begin{aligned}
    \!\hat{s}_i \!\!=\! & \mathop{\arg\max}\limits_{s_i} \mathrm{Pr}(o_i \mid s_i)\cdot \mathrm{Pr}(s_i \mid a_{i-1}, \hat{s}_{i-1}) \\
    = & \!\mathop{\arg\max}\limits_{k} \frac{\mathrm{Pr}(s_i\!\! =\!\! k \!\mid \!o_i)\mathrm{Pr}(o_i)}{\mathrm{Pr}(s_i = k)}\! \cdot \!\mathrm{Pr}(\!s_i\! =\! k \!\mid\! a_{i-1}\!,\! \hat{s}_{i-1}\!).
  \end{aligned}
\end{equation}
Here, Bayes’ theorem decomposes conditional probabilities into more tractable terms for estimation and learning.

We compute the probability term in Eq.~\eqref{eq:term_probability} using an imitation learning approach. Each training sample consists of a tuple $\langle$observation, state, action$\rangle$—a sequence of images (observations) labeled with the corresponding task stage (state) and expert action. The state labels are manually annotated.

The prior $\mathrm{Pr}(s_i = k)$ is estimated from the empirical state distribution in the training set. Since observations are always available, $\mathrm{Pr}(o_i)$ is set to 1. We then use supervised learning to estimate the conditional probabilities $\mathrm{Pr}(s_i = k \mid o_i)$ and $\mathrm{Pr}(s_i = k \mid a_{i-1}, \hat{s}_{i-1})$.

To estimate $\mathrm{Pr}(s_i = k \mid o_i)$, we use a neural network:
\begin{equation}~\label{eq:pred_s_1}
    \hat{\bm{s}}_i^{<o>} = \mathrm{NN}_o(o_i; \theta_o),
\end{equation}
where $\hat{\bm{s}}_i \in [0,1]^K$ is a $K$-dimensional probability vector, with its $k$-th element representing the estimated probability that $s_i = k$. The parameters $\theta_o$ are learnable. We define the one-hot vector $\bm{s}_i \in \{0,1\}^K$ as the ground truth for the state, where the $k$-th component is 1 if $s_i = k$. The model is trained to minimize the error between $\hat{\bm{s}}_i$ and $\bm{s}_i$.

Similarly, to estimate $\mathrm{Pr}({s}_i = k \mid a_{i-1}, \hat{\bm{s}}_{i-1})$, we use another network:
\begin{equation}~\label{eq:pred_s_2}
    \hat{\bm{s}}_i^{<a>} = \mathrm{NN}_a(a_{i-1}, \hat{\bm{s}}_{i-1}; \theta_a),
\end{equation}
where $\theta_a$ are learnable parameters of $\mathrm{NN}_a(\cdot)$. Substituting Eqs.~\eqref{eq:pred_s_1} and~\eqref{eq:pred_s_2} into Eq.~\eqref{eq:term_probability} yields the estimated state sequence $(\hat{\bm{s}}_1, \ldots, \hat{\bm{s}}_t)$.

\subsubsection{Optimizing Decision-Making Agent}
\label{subsubsec:agent}
Given the estimated states $(\hat{\bm{s}}_1, \ldots, \hat{\bm{s}}_t)$, the agent generates actions as { $\left(a_1 = \pi(o_1, \hat{\bm{s}}_1), \ldots, a_t = \pi(o_t, \hat{\bm{s}}_t)\right)$}.

At each $t$, the reward is defined as
\begin{equation}~\label{eq:reward-0}
    r_t(\theta_\pi) = - \Big\| \pi(o_t, \hat{\bm{s}}_t; \theta_\pi) - a'_t \Big\|,
\end{equation}
where $a'_t$ is the expert action. The objective is to optimize $\theta_\pi$ to maximize cumulative reward over the dataset:
\begin{equation}\label{eq:reward-1}
    \theta_\pi^{\ast} = \mathop{\arg\max}\limits_{\theta_\pi} 
    \sum_{t=1}^{T} r_t(\theta_\pi).
\end{equation}
where $r_t$ is the reward obtained at time step $t$, and $T$ denotes the total length of the training data.

\textbf{Remark:} In summary, implementing the HMDP framework requires two essential modules: A hidden state estimation module to infer the task stage from partial observations (Section \ref{subsubsec:infer}), and a decision-making agent to generate actions based on both the observation and the inferred state (Section \ref{subsubsec:agent}). In the next section, we describe how these modules are realized in our End-to-End imitation learning framework.
\section{HMDP in Real world Imitation Learning}
\label{sec:HMDP-Real-world}

In Section~\ref{subsec:HMDP}, we formulated the vision-based robotic manipulation problem as a Hidden Markov Decision Process (HMDP), which conceptually separates the task into two functional components: \textit{Hidden state estimation} and \textit{decision-making agent}. In this section, we instantiate these two components in practical imitation learning scenarios.

\begin{figure*}[htbp]
    \centering
    \includegraphics[width=1\textwidth]{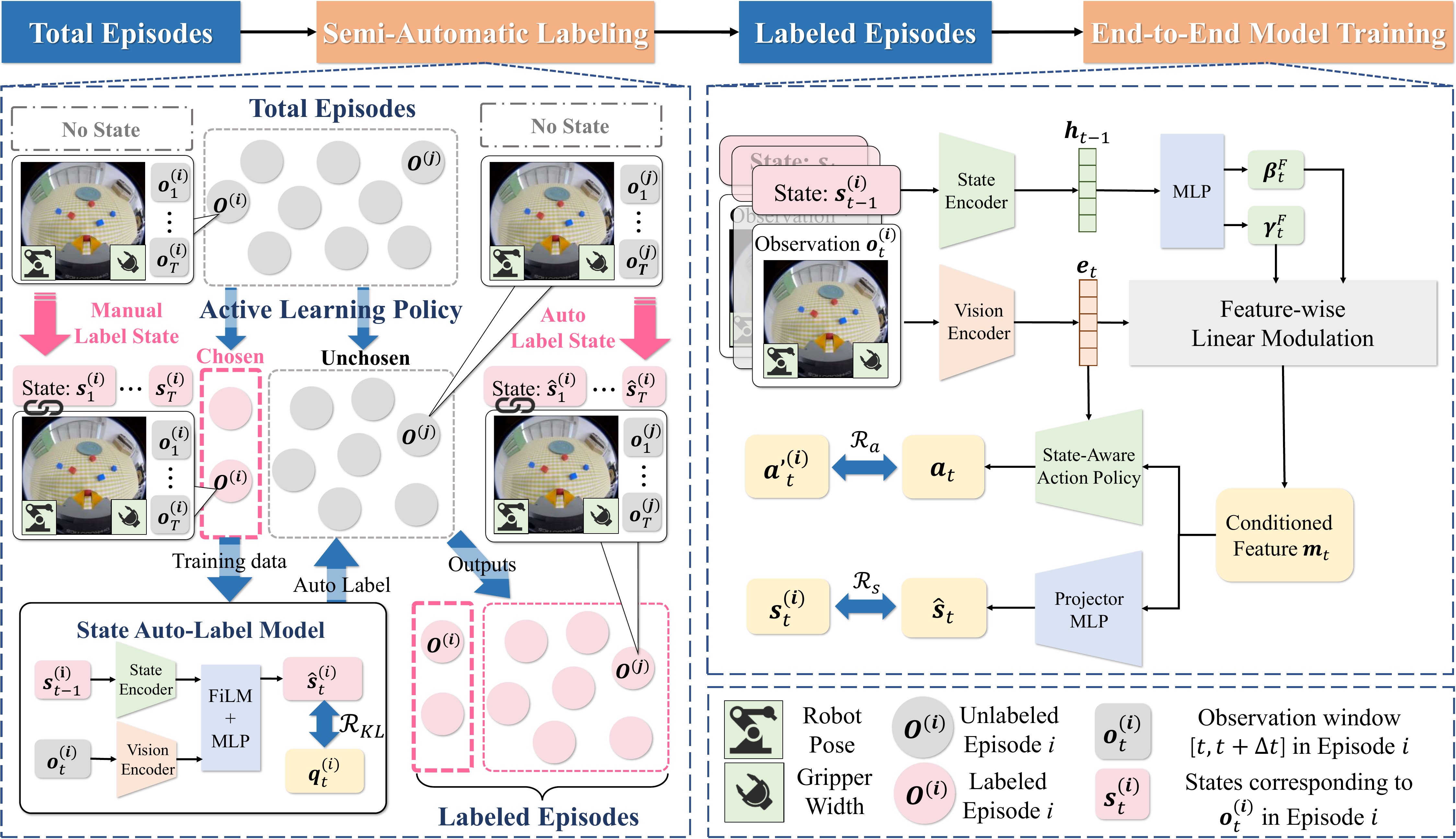}  
    \vspace{-0.5cm}
    \caption{The figure illustrates the process of active learning for selecting videos and automatically labeling states(left side), and our End-to-End model designed based on a Hidden Markov Decision Process(right side).
    }
    \vspace{-0.5cm}
    \label{fig:main}  
\end{figure*}

Section~\ref{subsec:state-transition-network} describes the \textit{hidden state estimation} module implemented with a neural network, in accordance with Eq.~\eqref{eq:pred_s_1} and Eq.~\eqref{eq:pred_s_2}. In Section~\ref{subsec:end-to-end-action-policy}, we realize a state-aware action policy, instantiated as the \textit{decision-making agent}'s ${\pi}$, which employs a diffusion-based method to generate actions, thus completing the HMDP pipeline from observation to action. These two modules are integrated into a unified End-to-End architecture, as illustrated on the right side of Fig.~\ref{fig:main}.

\subsection{State Transition Network}
\label{subsec:state-transition-network}

In this section, our objective is to implement the \emph{hidden state estimation module} within the Hidden Markov Decision Process (HMDP) framework. Guided by the theoretical formulations in Eq.~\eqref{eq:pred_s_1} and Eq.~\eqref{eq:pred_s_2}, the module infers the hidden state $\hat{\bm{s}}_t$ from the current observation $\bm{o}_t$ and the previous estimate $\hat{\bm{s}}_{t-1}$. Since these two equations capture complementary aspects of the hidden state, we unify them into a single network \footnote{Although Eq.~\eqref{eq:pred_s_2} includes the previous action $\bm{a}_{t-1}$, it is omitted here because the current observation $\bm{o}_t$ already contains proprioceptive information (e.g., joint states), which inherently reflects the previous action. Note that in the theoretical analysis we consider a simplified case; in practice, $\bm{o}$ and $\bm{a}$ are all vectors, hence the bold notation.}:
\begin{equation}
\hat{\bm{s}}_t = f(\hat{\bm{s}}_{t-1}, \bm{o}_t).
\end{equation}
Here, the previous state $\hat{\bm{s}}_{t-1}$ provides essential temporal context to disambiguate observations that look alike but correspond to different hidden states. By jointly encoding $\hat{\bm{s}}_{t-1}$ and $\bm{o}_t$, the network can better infer the correct hidden state.

The process involves two main steps: First, encoding both $\bm{o}_t$ and $\hat{\bm{s}}_{t-1}$ into feature representations through a vision encoder and a state encoder, respectively; second, fusing these features using Feature-wise Linear Modulation \cite{perez2018film}(FiLM) to enable the model to jointly leverage temporal context and visual information for accurate state prediction.

\subsubsection{Hidden State and Observation Encoding}

The hidden state $\hat{\bm{s}}_{t-1}$ is first discretized into a one-hot vector during labeling based on the current task stage. This one-hot vector is then mapped to an embedding $\bm{h}_{t-1}$ through a \emph{state encoder}, which enhances its representational capacity. Simultaneously, we use a \emph{vision encoder} to transform the observation $\bm{o}_t$ into an embedding $\bm{e}_t \in \mathbb{R}^{D_e}$, as described in Eq.~\eqref{eq:encode}. The vision encoder extracts high-level visual features that capture the current state of the environment, making them useful for subsequent decision-making.
\begin{equation}\label{eq:encode}
\bm{e}_t = \mathrm{Enc}_{\text{vision}}(\bm{o}_t), \quad \bm{h}_{t-1} = \mathrm{Enc}_{\text{state}}(\hat{\bm{s}}_{t-1})
\end{equation}

\subsubsection{Feature-wise Linear Modulation}

In order to fuse the two modalities—visual and state embeddings—in network $f$, we employ FiLM~\cite{perez2018film}. The state embedding $\bm{h}_{t-1}$, being low-dimensional, is first projected into the same feature space as the visual encoder output using a multi-layer perceptron:
\begin{equation}\label{eq:E0}
\bm{\gamma}_{t}^{\text{F}}, \bm{\beta}_{t}^{\text{F}} = \mathrm{MLP}(\bm{h}_{t-1}),
\end{equation}
where $\bm{\gamma}_{t}^{\text{F}}, \bm{\beta}_{t}^{\text{F}} \in \mathbb{R}^{D_e}$ are the FiLM modulation parameters that adaptively scale and shift the visual features at time $t$. Next, we use these parameters to modulate the visual feature $\bm{e}_t$. 
The fused representation $\bm{m}_t$ is computed as:
\begin{equation}\label{eq:mt}
\bm{m}_t = \lambda_1 \left(\bm{e}_t \odot \bm{\gamma}_{t}^{\text{F}} + \bm{\beta}_{t}^{\text{F}}\right) + (1 - \lambda_1) \bm{e}_t,
\end{equation}
where $\odot$ denotes element-wise multiplication, and $\lambda_1$ is a regularization weight that controls the impact of the state modulation. This fusion mechanism enables the model to disambiguate visual observations by incorporating the contextual information provided by the hidden state $\bm{h}_{t-1}$, thereby resolving any confusion that might arise from visually similar inputs. Finally, the fused feature $\bm{m}_t$ is passed through a projector MLP and a softmax layer to generate the predicted state distribution $\hat{\bm{s}}_t$:
\begin{equation}\label{eq:H2}
\hat{\bm{s}}_t = \mathrm{SoftMax}(\mathrm{MLP_{proj}}(\bm{m}_t)).
\end{equation}

This output represents the model's predicted hidden state for the current task, which is used to inform action generation in downstream modules. To train the state transition network $f$, we treat the negative cross-entropy between the predicted state $\hat{\bm{s}}_t$ and the ground truth $\bm{s}_t$ as the reward:
\begin{equation}\label{eq:L1}
\mathcal{R}_s = \sum_{i=1}^{d_s} \bm{s}_{t,i} \cdot \log \hat{\bm{s}}_{t,i},
\end{equation}
where $d_s$ denotes the dimension of the discrete state, 
$\bm{s}_{t,i}$ and $\hat{\bm{s}}_{t,i}$ represent the $i$-th components of the ground-truth one-hot state vector $\bm{s}_t$ and the predicted categorical distribution $\hat{\bm{s}}_t$, respectively. During training, $\bm{s}_t$ is manually annotated to provide supervision for state prediction.


\subsection{State-Aware Action Policy}
\label{subsec:end-to-end-action-policy}

In Section~\ref{subsec:state-transition-network}, we introduced the state transition network, which infers the hidden state $\hat{\bm{s}}_t$ at each time step. 
Here, we construct the state-aware action policy ${\pi}$ based on both the observation and state information:
\begin{equation}\label{eq:a=pi}
    \bm{a}_t = {\pi}(\bm{o}_t, \hat{\bm{s}}_{t}),
\end{equation}
where $\bm{a}_t$ denotes the action at time step $t$. We implement ${\pi}$ following the Diffusion Policy framework~\cite{chi2023diffusion}.

\subsubsection{Diffusion Policy}

\textbf{Action Generation.} Diffusion Policy (DP)~\cite{chi2023diffusion} formulates action generation as a conditional denoising diffusion process, conditioned on the visual feature $\bm{e}_t$. The policy initializes an action sequence $\bm{a}^K_t \sim \mathcal{N}(0,I)$ and iteratively refines it by the reverse update:
\begin{equation}\label{eq:dp_denoise}
    \bm{a}^{k-1}_t = \alpha_k \Big( \bm{a}^k_t - \gamma_k \, \epsilon_\theta(\bm{e}_t, \bm{a}^k_t, k) \Big) 
              + \sigma_k \, \mathcal{N}(0,I),
\end{equation}
where $\epsilon_\theta$ is the noise prediction network, and $\alpha_k$, $\gamma_k$, and $\sigma_k$ are schedule-dependent coefficients defined following DDPM~\cite{Ho2020DDPM}. 
Here $k$ denotes the diffusion step index, which decreases from $K$ (initial Gaussian noise) to $0$ (clean action sequence). After $K$ denoising steps, the final sequence $\bm{a}^0_t$ is obtained, from which the action is executed at time $t$.

\textbf{Training objective.} To train the denoising network $\epsilon_\theta$, DP randomly samples a clean action sequence $\bm{a}^0_t$ from the dataset and applies the forward diffusion process to obtain its noisy counterpart at step $k$:
\begin{equation}
    \bm{a}^k_t = \sqrt{\bar{\alpha}_k}\, \bm{a}^0_t + \sqrt{1-\bar{\alpha}_k}\,\epsilon, 
    \quad \epsilon \sim \mathcal{N}(0,I),
\end{equation}
where $\bar{\alpha}_k$ is the cumulative product of the noise scheduler coefficients up to step $k$~\cite{Ho2020DDPM}. 
The network is trained to predict the injected noise $\epsilon$ from the noisy action sequence $\bm{a}^k_t$, conditioned on the observation embedding $\bm{e}_t$, and the prediction accuracy is treated as an action reward:
\begin{equation}\label{eq:L2}
\mathcal{R}_{\text{a}} = - \frac{1}{d_a} \big\| \epsilon - \epsilon_\theta(\bm{e}_t, \bm{a}^k_t, k) \big\|_2^2,
\end{equation}
where $d_a$ is the dimensionality of the action vector $\bm{a}_t$.

\subsubsection{State-aware Action}
As discussed earlier, resolving state ambiguity requires not only visual input but also state information. Therefore, we replace the original conditioning signal $\bm{e}_t$ in Eq.~\eqref{eq:dp_denoise} with the fused representation $\bm{m}_t$ from Eq.~\eqref{eq:mt}, which integrates both the current visual observation and the inferred state information\footnote{As in Eq.~\eqref{eq:a=pi}, the policy depends on $\hat{\bm{s}}_t$ and $\bm{o}_t$. Since $\bm{m}_t$ is derived from $\bm{o}_t$ and $\hat{\bm{s}}_{t-1}$, it preserves visual and state information. Moreover, $\hat{\bm{s}}_t$ is inferred from $\bm{m}_t$, making their features highly aligned. Thus, $\bm{m}_t$ alone suffices as a compact, state-aware substitute.}:
\begin{equation}\label{eq:sage_denoise}
    \bm{a}^{k-1}_t = \alpha_k \Big( \bm{a}^k_t - \gamma_k \, \epsilon_\theta(\bm{m}_t, \bm{a}^k_t, k) \Big) 
              + \sigma_k \, \mathcal{N}(0,I).
\end{equation}

While this provides a unified input signal, it may still be insufficient for action prediction in certain cases. 
For example, in \textit{push buttons} tasks (Fig.~\ref{fig:introduction}), the associated actions share a similar downward gripper motion, with the primary difference being the lateral displacement toward the correct button. 
Such subtle variations can be difficult to capture if the conditioning signal is not strong enough. 
Therefore, the policy should strengthen the influence of the state on action generation, ensuring that identical observations under different states can yield distinct actions when required.

Building upon the concept of classifier-free guidance (CFG)~\cite{ho2022classifier}, we extend the standard action policy to incorporate state-awareness, explicitly addressing this challenge. Specifically, we train the denoising model $\epsilon_\theta$ with two types of conditional inputs by randomly dropping the state signal during training:

\begin{equation}
\epsilon_\theta =
\begin{cases}
\epsilon_\theta(\bm{e}_t, \bm{a}^k_t, k), & \text{with prob. } p_{\text{drop}}, \\[4pt]
\epsilon_\theta(\bm{m}_t, \bm{a}^k_t, k), & \text{with prob. } 1-p_{\text{drop}},
\end{cases}
\label{eq:cfg-train}
\end{equation}
where $\bm{e}_t$ is the observation embedding from Eq.~\eqref{eq:encode}, $\bm{m}_t$ is the fused state-aware embedding from Eq.~\eqref{eq:mt}, and $\bm{a}^k_t$ is the noisy action sequence at step $k$. Here $p_{\text{drop}} \in (0,1)$ is a hyperparameter that controls the probability of dropping the state signal during training. This enables the network to learn both \emph{state-agnostic} and \emph{state-aware} prediction modes. 

At inference, the two modes are combined via a guidance scale $w>0$ to amplify the causal role of the state: 
\begin{equation} \label{eq:cfg-infer} \small
\tilde{\epsilon}_\theta(\bm{m}_t\!,\bm{a}^k_t\!,k)
\!=\! \epsilon_\theta(\bm{m}_t\!,\bm{a}^k_t\!,k)
\!+\! w\big(\epsilon_\theta(\bm{m}_t\!,\bm{a}^k_t\!,k) 
\!-\! \epsilon_\theta(\bm{e}_t\!,\bm{a}^k_t\!,k)\big).
\end{equation}
In essence, the guided prediction $\tilde{\epsilon}_\theta$ consists of the base term plus a residual correction, which amplifies the contribution of state-aware features relative to state-agnostic ones. During inference, we replace $\epsilon_\theta$ with $\tilde{\epsilon}_\theta$ in the reverse denoising step (Eq.~\eqref{eq:sage_denoise}), ensuring that the same observation $\bm{o}_t$ under different hidden states can lead to distinct action predictions. 

In this way, our action decision policy extends the standard Diffusion Policy~\cite{chi2023diffusion} by incorporating classifier-free guidance with state-aware conditioning. 
By jointly training on both state-agnostic and state-aware inputs and applying guided inference, the policy ${\pi}$ achieves stronger state-to-action coupling, which is critical for resolving state ambiguity.

\subsection{End to End Model Training}
\label{subsec:end-to-end}
This section outlines the smooth flow of the labeling and training process. Since the model introduces an additional state variable $\bm{s}_t$, the training procedure in Eq.~\eqref{eq:L1} requires corresponding state supervision signals. Therefore, the entire training pipeline begins with manually annotating the collected dataset, followed by model training.

\subsubsection{Dataset Preparation}
In the data collection phase, each demonstration consists of a sequence of observations and corresponding actions:
\begin{equation}\label{eq:def-active-0}
\bm{O}^{(n)} = \{\bm{o}_t^{(n)}\}_{t=1}^{T_n}, \quad
\bm{A}^{(n)} = \{{\bm{a}'}_t^{(n)}\}_{t=1}^{T_n},
\end{equation}
where the superscript $^{(n)}$ denotes the $n$-th episode, and $T_n$ denotes the length of the $n$-th episode. We collect $N$ episodes for training, and the full demonstration dataset is denoted as:
\begin{equation}\label{eq:def-active-1}
\mathcal{B} = \left\{ \left( \bm{o}_t^{(n)}, {\bm{a}'}_t^{(n)} \right)_{t=1}^{T_n} \right\}_{n=1}^N.
\end{equation}

To enable state-aware learning, we annotate each frame with a hidden state by reviewing video replays, thus obtaining the state sequence corresponding to Eq.~\eqref{eq:def-active-0}:
\begin{equation}\label{eq:def-active-2}
\bm{S}^{(n)} = \{\bm{s}_t^{(n)}\}_{t=1}^{T_n}, 
\quad \bm{s}_t^{(n)} \in \{0,1\}^{d_s},
\end{equation}
where each $\bm{s}_t^{(n)}$ is a one-hot vector indicating the high-level stage the agent is in at time $t$. The resulting labeled dataset becomes:
\begin{equation}\label{eq:def-active-3}
\mathcal{B}_{\text{labeled}} = \left\{ \left( \bm{o}_t^{(n)}, {\bm{a}'}_t^{(n)}, \bm{s}_t^{(n)} \right)_{t=1}^{T_n} \right\}_{n=1}^N.
\end{equation}

\subsubsection{Training Process}
Next, we proceed with training the model using the dataset from Eq.~\eqref{eq:def-active-3}, applying the models introduced in previous sections. In this dataset, the state labels $\bm{s}_t$ provide the necessary supervisory signal for the state transition network discussed in Section~\ref{subsec:state-transition-network}, while the action labels $\bm{a}_t'$ serve as the supervisory signal for the state-aware action policy model outlined in Section~\ref{subsec:end-to-end-action-policy}.

The total reward combines $\mathcal{R}_s$ and $\mathcal{R}_a$, as defined in Eq.~\eqref{eq:L1} and Eq.~\eqref{eq:L2}, respectively. Here, $\mathcal{R}_s$ quantifies the reward for accurately predicting the hidden state, while $\mathcal{R}_a$ reflects the reward associated with action generation. During training, the action sequences are sampled according to the state-agnostic and state-aware probabilities specified in Eq.~\eqref{eq:cfg-train}, enabling the model to learn both conditional modes. The total reward is given by:
\begin{equation}~\label{eq:L3}
\mathcal{R} = \mathcal{R}_s + \lambda_2 \mathcal{R}_a,
\end{equation}
where $\lambda_2 \in \mathbb{R}$ is a weighting factor that balances the contributions of state prediction and action decision in the reward function.

\section{Learning to Annotate States Efficiently}
\label{sec:auto_tag}
\label{subsec:auto_tag}

In Section~\ref{sec:HMDP-Real-world}, we proposed a Hidden Markov Decision Process (HMDP)-based modeling framework, where task stages are introduced as hidden state variables to impose structural priors over the observation sequence. However, the inclusion of hidden state variables $\bm{s}_t$ in Eq.~\eqref{eq:def-active-3} introduces an additional annotation burden during training, leading to two key issues:

\begin{algorithm}[t]
\caption{Semi-Automatic Labeling}
\label{alg:semi_auto_tag}
\begin{algorithmic}[1]
\STATE \textbf{Input:} Full raw dataset $\mathcal{B}$
\STATE $\mathcal{C} \leftarrow \texttt{SelectSubset}(\mathcal{B})$\label{line:selectsubset} \hfill // Active Learning

\STATE $\mathcal{C}_{\text{labeled}} \leftarrow  \texttt{ManualLabel}(\mathcal{C})$ \hfill // Manual annotation

\STATE Train \textit{State Auto-Label Model} $H$ on dataset $\mathcal{C}_{\text{labeled}}$ via Eq.~\eqref{eq:L1} \label{line:reward}

\STATE $(\mathcal{B} - \mathcal{C})_{\text{auto}} \leftarrow H(\mathcal{B} - \mathcal{C})$ 
\hfill // Automatically label the remaining subset $\mathcal{B} - \mathcal{C}$ using $H$

\STATE $\mathcal{B}_{\text{labeled}} \leftarrow \mathcal{C}_{\text{labeled}} \cup (\mathcal{B} - \mathcal{C})_{\text{auto}}$

\STATE \textbf{Output:} Labeled dataset $\mathcal{B}_{\text{labeled}}$
\end{algorithmic}
\end{algorithm}

\begin{enumerate}[label=\textbf{Q\arabic*}, leftmargin=2.5em]
    \item \label{q1}
    Each observation frame requires frame-level state labels, resulting in significant manual annotation effort. Is there a way to reduce the manual annotation cost?
    \item \label{q2}
    State transitions occur sparsely within individual episodes, making it hard for the model to learn transition boundaries from limited labeled data. How can we address this learning difficulty?
\end{enumerate}

To address these issues, we propose a hybrid annotation strategy that combines \textit{active learning} and \textit{soft label interpolation}, as detailed in Sections~\ref{subsec:active_tag} and~\ref{subsec:soft_label}, respectively.

In Section~\ref{subsec:active_tag}, we address the annotation burden in~\ref{q1} by employing an active learning strategy. As defined in Eq.~\eqref{eq:def-active-1}, let $\mathcal{B}$ denote the full set of expert demonstration episodes. Our objective is to identify a representative subset $\mathcal{C} \subset \mathcal{B}$ for manual state annotation. A \textit{State Auto-Label Model} is first trained on a small manually labeled subset $\mathcal{C}$ of the dataset, where $\mathcal{C} \ll \mathcal{B}$. It is then used to automatically annotate the much larger remaining set $\mathcal{B} - \mathcal{C}$, thereby eliminating the need for extensive manual labeling and significantly reducing the overall annotation workload.
The whole process is shown in Algorithm~\ref{alg:semi_auto_tag}.

To further address the challenge in~\ref{q2}, we introduce a soft label enhancement mechanism in Section~\ref{subsec:soft_label}, which smooths the manually labeled states in $\mathcal{C}$ to improve supervision quality and generalization. The overall pipeline for solving ~\ref{q1} and ~\ref{q2} is illustrated on the left side of Fig.~\ref{fig:main}.

\subsection{Active Learning}
\label{subsec:active_tag}

Active learning is grounded in the principle that not all samples contribute equally to model training; by selectively labeling the most informative or representative data, one can significantly reduce annotation cost while preserving model performance. In the following, we discuss the strategy for selecting the optimal subset to implement the \texttt{SelectSubset} function in Algorithm~\ref{alg:semi_auto_tag}, line~\ref{line:selectsubset}, thereby maximizing the effectiveness of the manual annotation budget.

\begin{algorithm}[t]
\caption{Greedy Selection of Subset $\mathcal{C}$}
\label{alg:alg1}
\begin{algorithmic}[1]
\STATE \textbf{Input:} Dataset $\mathcal{B}$, selection budget $b$
\STATE $\mathcal{C} \leftarrow \emptyset$
\STATE Train vision encoder using Eq.~\eqref{eq:L2} (omitting stage label $\bm{s}_t$) to obtain parameters $\theta$
\FOR{each episode $m \in \mathcal{B}$}
    \FOR{each timestep $t$ in episode $m$}
        \STATE $\bm{e}_t^{(m)} \leftarrow \text{Enc}_\text{vision}(\bm{o}_t^{(m)}; \theta)$ \hfill // Eq.~\eqref{eq:encode}
    \ENDFOR
\ENDFOR
\WHILE{$|\mathcal{C}| < b$}
    \STATE $i^\ast \leftarrow \mathop{\arg\!\min}\limits_{i \in \mathcal{B}-\mathcal{C}} \psi(\mathcal{C} \cup \{i\})$ \hfill // Eq.~\eqref{eq:S3}
    \STATE $\mathcal{C} \leftarrow \mathcal{C} \cup \{i^\ast\}$
\ENDWHILE
\STATE \textbf{Return} $\mathcal{C}$
\end{algorithmic}
\end{algorithm}

\subsubsection{Representativeness Estimation}
We aim to select a subset { $\mathcal{C}$} that adequately covers the distribution of the full dataset { $\mathcal{B}$} in the feature space, thus ensuring good generalization. Inspired by the Core-set approach~\cite{sener2018active}, we evaluate the representativeness of samples based on feature similarity.

Formally, given any two observations {$\bm{o}^{(n)}_\tau$} and {$\bm{o}^{(m)}_t$} from episodes {$n$}  and {$m$}, respectively, we extract their feature representations  {$\bm{e}^{(n)}_\tau$}  and  {$\bm{e}^{(m)}_t$}  using Eq.~\eqref{eq:encode}. The cosine distance between these vectors is defined as:
\begin{equation}\label{eq:S1}
	d\left(\bm{e}_t^{(m)},\bm{e}_\tau^{(n)} \right)=1-\frac{\left<\bm{e}_t^{(m)},\bm{e}_\tau^{(n)} \right>}{\left|\left|\bm{e}_t^{(m)} \right|\right|\cdot\left|\left|\bm{e}_\tau^{(n)} \right|\right|},
\end{equation}
where {$\left<\cdot,\cdot\right>$} denotes the dot product and {$\left||\cdot\right||$} is the {$l_2$} norm. Function $\small d\left(\bm{e}_t^{(m)}, \bm{e}_\tau^{(n)} \right)$ describes the similarity between any two observations at the frame level.

We further define the representational ability of episode { $\bm{O}^{(n)}$} for episode { $\bm{O}^{(m)}$} as:
\begin{equation}\label{eq:S2}
\Delta\left(\bm{O}^{(m)}, \bm{O}^{(n)}\right) = \max_{t} \min_{\tau} d\left( \bm{e}_t^{(m)}, \bm{e}_\tau^{(n)} \right),
\end{equation}
where, for each representation $\bm{e}_t^{(m)}$ in $\bm{O}^{(m)}$, we find its nearest representation $\bm{e}_\tau^{(n)}$ in $\bm{O}^{(n)}$, and then select the maximum value among all the minimum distances (i.e., $\max_t \min_\tau d(\cdot)$). The metric $\Delta\left(\bm{O}^{(m)}, \bm{O}^{(n)}\right)$ thus measures how well $\bm{O}^{(n)}$ can ``cover'' $\bm{O}^{(m)}$ in the feature space, which is episode-level. 

Therefore, we aim to ensure that the episodes in the subset $\mathcal{C}$ can ``cover'' the episodes in $\mathcal{B} - \mathcal{C}$ as much as possible. To this end, we define a function $\psi$ to quantify the  coverage capability.
\begin{equation}\label{eq:S3}
\psi\left(\mathcal{C}\right)=\max_{m\in\mathcal{B-C}}\min_{n\in\mathcal{C}}\Delta\left(\bm{O}^{(m)},\bm{O}^{(n)}\right).
\end{equation}
Here, $\psi(\mathcal{C})$ computes, for each episode $m$ in $\mathcal{B} - \mathcal{C}$, the minimum feature-space distance to any episode $n$ in $\mathcal{C}$, and then takes the maximum over all such episodes in $\mathcal{B} - \mathcal{C}$. A smaller value of $\psi(\mathcal{C})$ indicates that even the most difficult-to-cover episode in $\mathcal{B} - \mathcal{C}$ has a close representative in $\mathcal{C}$, implying stronger overall coverage.

\subsubsection{Greedy Selection for Auto-Labeling}

Our goal is to select a subset $\mathcal{C}$ that minimizes $\psi(\mathcal{C})$. 
However, this corresponds to the NP-hard $k$-center problem. 
To address this, we adopt a greedy algorithm that iteratively adds samples to $\mathcal{C}$, each time selecting the sample that yields the largest reduction in $\psi(\mathcal{C})$, until the size of $\mathcal{C}$ reaches a predefined budget $b$, as shown in Alg.~\ref{alg:alg1}. 
In fact, Alg.~\ref{alg:alg1} exactly implements the \texttt{SelectSubset} function required in Alg.~\ref{alg:semi_auto_tag}, line~\ref{line:selectsubset}.

Once $\mathcal{C}$ is obtained, we manually annotate its states to obtain $\mathcal{C}_{\text{labeled}}$, which has the same structure as the dataset defined in Eq.~\eqref{eq:def-active-3}, as illustrated in Alg.~\ref{alg:semi_auto_tag}. 
$\mathcal{C}_{\text{labeled}}$ is then used to train the \textit{State Auto-Label Model}. 
This model is subsequently employed to automatically annotate the remaining unlabeled state sequences. This completes the semi-automatic data annotation process based on active learning, resulting in a substantial reduction in manual labeling costs.

\vspace{-0.5cm}
\subsection{Regularizing State Transitions via Soft States}
\label{subsec:soft_label}

\begin{figure}[t]
    \centering
    \includegraphics[width=0.45\textwidth]{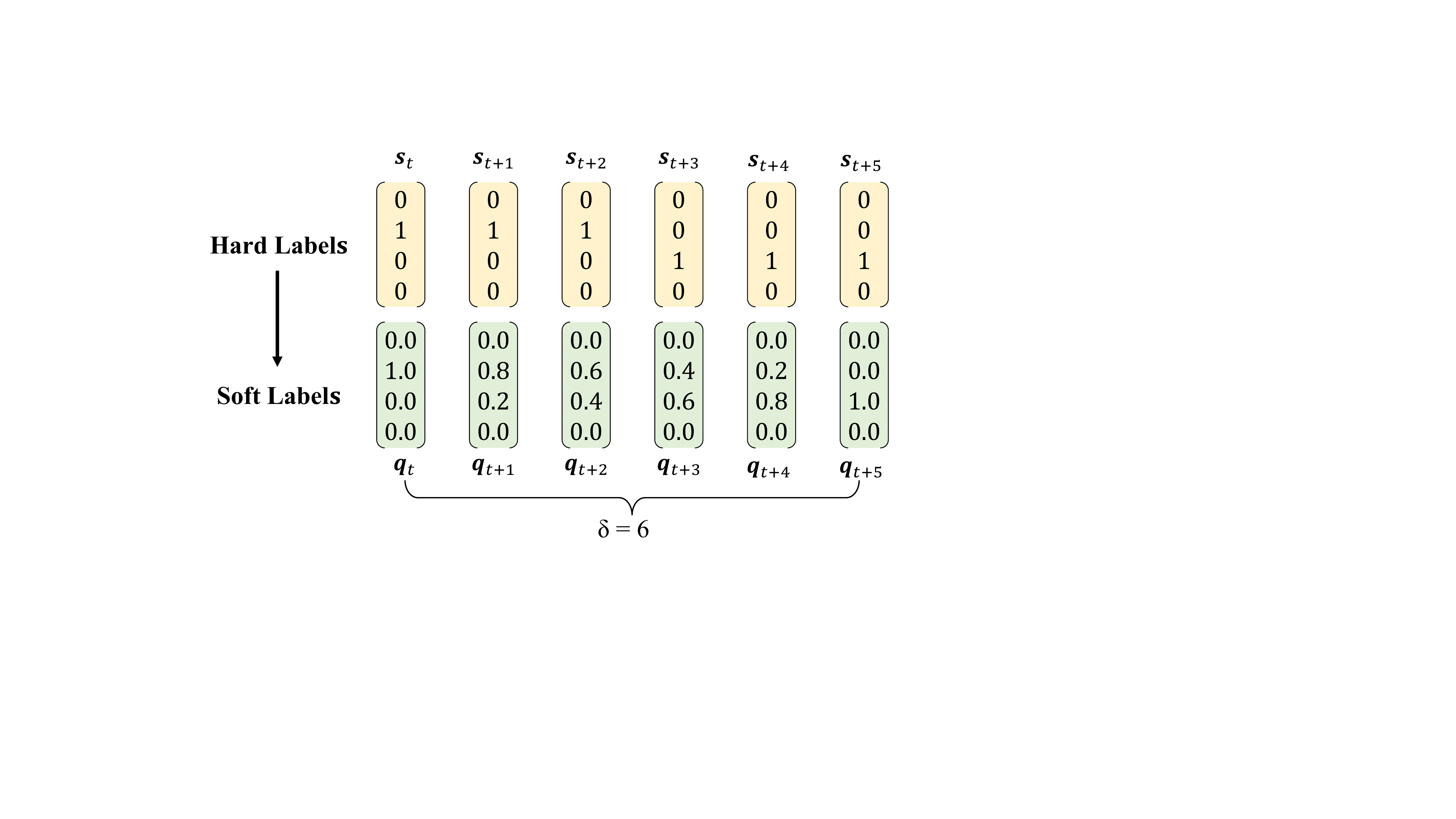}
    \caption{
An example of soft label generation from hard labels using Eq.~\eqref{eq:soft-label}, with a transition window size of \(\delta = 6\) for illustration. Note that \(\delta = 6\) is used here only for visualization and does not reflect the actual value used in experiments.
}
    \vspace{-0.5cm}
    \label{fig:label}
\end{figure}

Since the original state labels $\bm{s}_t$ are discrete one-hot vectors, 
the model can become overconfident in prolonged periods of identical states, 
making it difficult to learn transitions effectively. 
Inspired by Label Smoothing Regularization~\cite{rethink}, 
we introduce a soft state label transformation $\bm{q}_t$ that distributes 
probability mass around state transitions, encouraging the model to remain 
sensitive to changes while mitigating overconfidence:
\begin{equation}
\label{eq:soft-label}
\bm{q}_t = 
\begin{cases}
\left( \! \frac{\delta - t + t^*}{2\delta} \! \right) \bm{s}_{t^*\!-\!1}\! +\! \frac{t - t^* + \delta}{2\delta} \bm{s}_{t^*},\!\!\! & t^*\!\! -\! \delta \!\leq t\! \leq t^*\!\! +\! \delta \\
\bm{s}_t, & \text{otherwise}
\end{cases}
\end{equation}
where $t^*$ is the timestamp such that $\bm{s}_{t^*-1}\neq\bm{s}_{t^*}$, and $\delta \in \mathbb{N}$ is a hyperparameter controlling the transition interval. An example of this transformation with $\delta=6$ is shown in Fig.~\ref{fig:label}, which provides a clearer understanding of the approach.

Before training the \textit{State Auto-Label Model}, we first obtain the soft state labels $\bm{q}_t$. 
To encourage more effective learning, in Algorithm~\ref{alg:semi_auto_tag} 
(line~\ref{line:reward}) we replace the reward originally defined by Eq.~\eqref{eq:L1} 
with a KL-divergence-based reward:
\begin{equation}\label{eq:loss-AL-kl}
\mathcal{R}_{\mathrm{KL}} = -\sum_{i=1}^{d_s} 
\bm{q}_{t,i} \cdot \left( \log \bm{q}_{t,i} - \log \hat{\bm{s}}_{t,i} \right),
\end{equation}
where $\bm{q}_{t,i}$ and $\hat{\bm{s}}_{t,i}$ are the {\small $i$}-th components of ground truth and prediction, respectively. All other components of the model remain unchanged.

By introducing the soft label approach, we can address the issue \ref{q2} encountered in the few-shot learning setting of Section~\ref{subsec:active_tag}, thereby improving the overall accuracy of the entire pipeline in Algorithm~\ref{alg:semi_auto_tag}.

\section{Real World Experiment}
\subsection{Experiment Setup}
\begin{figure}[!t]
    \centering
    \includegraphics[width=0.5\textwidth]{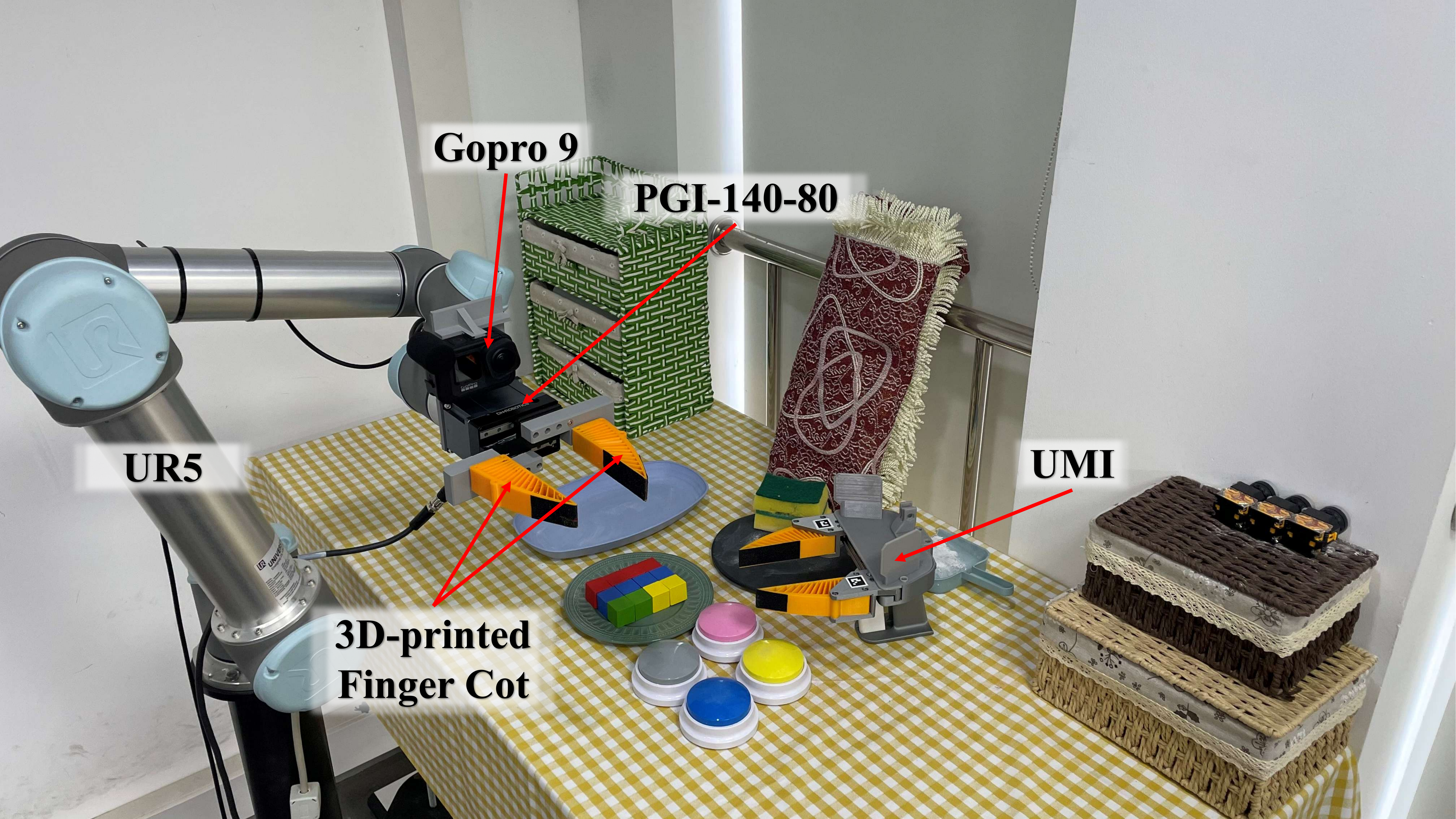}
    \vspace{-0.5cm}
    \caption{Experimental setup with task props and workspace.}
    \label{fig:workspace}
    \vspace{-0.5cm}
\end{figure}

\textbf{Workspace.}
Experimental workspace setup is shown in Fig.\ref{fig:workspace}. All experimental data were collected using the Universal Manipulation Interface(UMI) \cite{DBLP:conf/rss/ChiXPCB0TS24}. The workspace consists of a UR5 robotic arm, a Gopro 9 camera mounted on the wrist to provide an egocentric (eye-in-hand) view, and a PGI-140-80 industrial gripper. The Gopro 9 serves as the sole source of visual input throughout all tasks. 3D-printed finger cots are attached to the end-effectors for grasping and manipulation. Objects with diverse shapes, colors, and textures are placed in the workspace to enable various manipulation scenarios. For each task, we recorded 150–200 episodes as training datasets.

\begin{figure*}[t]
    \centering
    \includegraphics[width=1\textwidth]{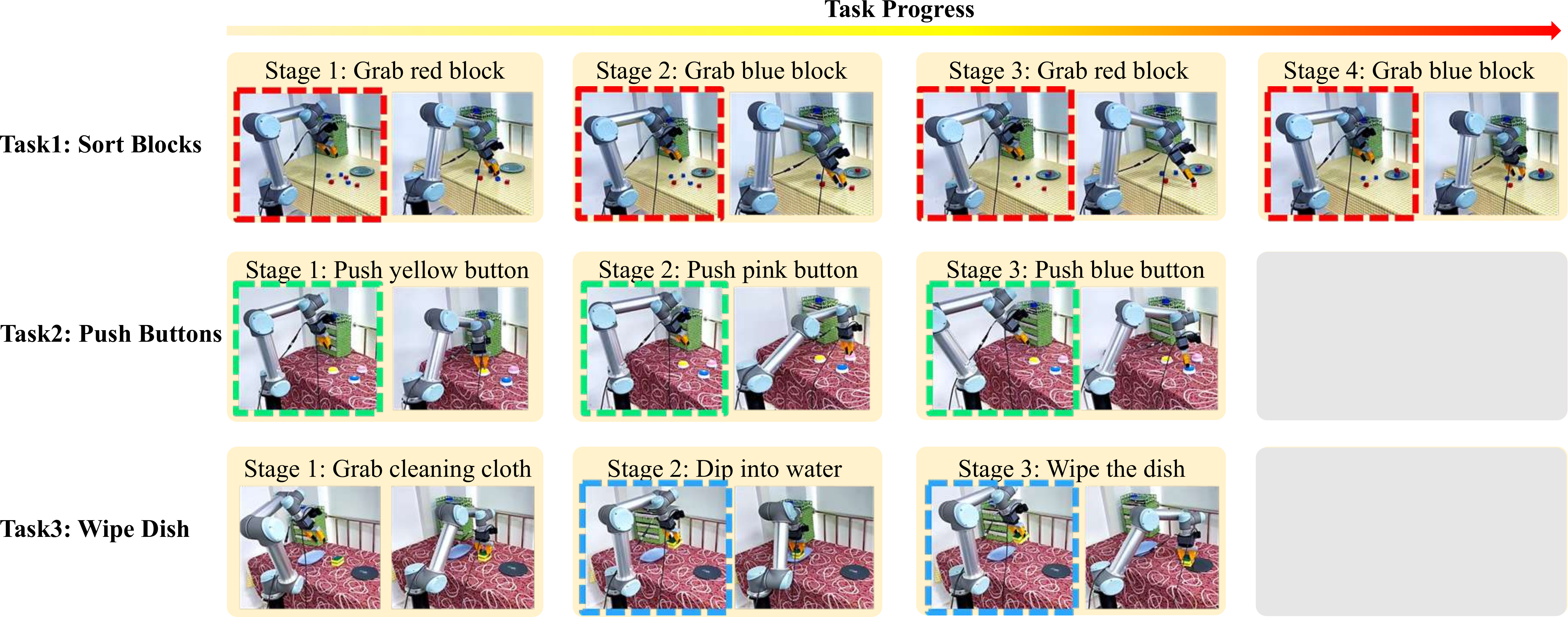}
    \vspace{-0.5cm}
    \caption{Three tasks design and the requirements of each one. \textbf{Row1: Sort blocks.} The workspace contains a tray and multiple blocks. The robotic arm is required to place four blocks into the tray in the strict sequence: red $\rightarrow$ blue $\rightarrow$ red $\rightarrow$ blue. \textbf{Row2: Push buttons.} The workspace contains three colored buttons: Yellow, pink, and blue. The robotic arm must press them in the specified order: Yellow $\rightarrow$ pink $\rightarrow$ blue. \textbf{Row3: Wipe Dish.} The workspace includes a water basin, a dry cleaning cloth, and a dish with visible stains. The robotic arm must complete the following sequence: grab the cloth $\rightarrow$ dip into water $\rightarrow$ wipe the dish. The dashed box indicates stages with state ambiguity within the task.
    }
    \label{fig:Task-Design}  
    \vspace{-0.5cm}
\end{figure*}

\textbf{Baselines.} In our experiments, we consider three categories of baselines: Visuomotor policies, memory-based models, and hierarchical models.

$\bullet$ \emph{Diffusion Policy}~\cite{chi2023diffusion}: Diffusion Policy (DP) is a state-of-the-art visuomotor policy that employs diffusion processes to predict robot actions. It serves as a representative model for vision-driven approaches in robotic manipulation tasks.

$\bullet$ \emph{BC-RNN}~\cite{mandlekar2022matters}: BC-RNN is a representative memory-based model that leverages recurrent neural networks to capture long-term dependencies in observations. Among several alternatives, including HBC~\cite{Hierarchical_Behavioral_Cloning(HBC)}, BCQ~\cite{BCQ}, CQL~\cite{Q-learning}, and IRIS~\cite{IRIS}, BC-RNN has demonstrated the best performance~\cite{mandlekar2022matters} and is therefore selected as our baseline.

$\bullet$ \emph{Hierarchical Models}: Hierarchical models decompose tasks into high-level planning and low-level motor execution. We adapt and reproduce the core ideas from hierarchical control literature~\cite{luo2024multistage,HVIL,gupta2019relay}, following design principles rooted in the modularity and reactivity emphasized by Behavior Trees~\cite{Behavior-Tree}.

\textbf{Task Design.}
We designed three types of robot manipulation tasks with \textit{state ambiguity}: \textit{Sort Blocks}, \textit{Push Buttons}, and \textit{Wipe Dish}. Examples of each task along with their corresponding specifications are illustrated in Fig.~\ref{fig:Task-Design}.

$\bullet$ \emph{Sort Blocks}. In this task, each stage exhibits \textit{state ambiguity}. At any given time, the observation consists of a cluttered arrangement of blocks in various colors, while the target color to be grasped differs across stages. This is illustrated in Fig.~\ref{fig:Task-Design}, Row~1, where stages with \textit{state ambiguity} are highlighted by red dashed boxes.

$\bullet$ \emph{Push Buttons}. Similarly, in the \textit{Push Buttons} task, each stage also suffers from \textit{state ambiguity}. The buttons look nearly identical before and after being pressed, so at any given time the observation contains several visually similar buttons. This is illustrated in Fig.~\ref{fig:Task-Design}, Row~2, where stages with \textit{state ambiguity} are highlighted by green dashed boxes.

$\bullet$ \emph{Wipe Dish}. This task presents \textit{state ambiguity} only at stages 2 and 3, as highlighted by the blue dashed box in Fig.~\ref{fig:Task-Design}. At these stages, the observation consists of the gripper holding a cloth, and the dry and wet cloths are nearly indistinguishable in appearance, which leads to \textit{state ambiguity}.

\begin{figure}[t]
    \centering
    \includegraphics[width=0.5\textwidth]{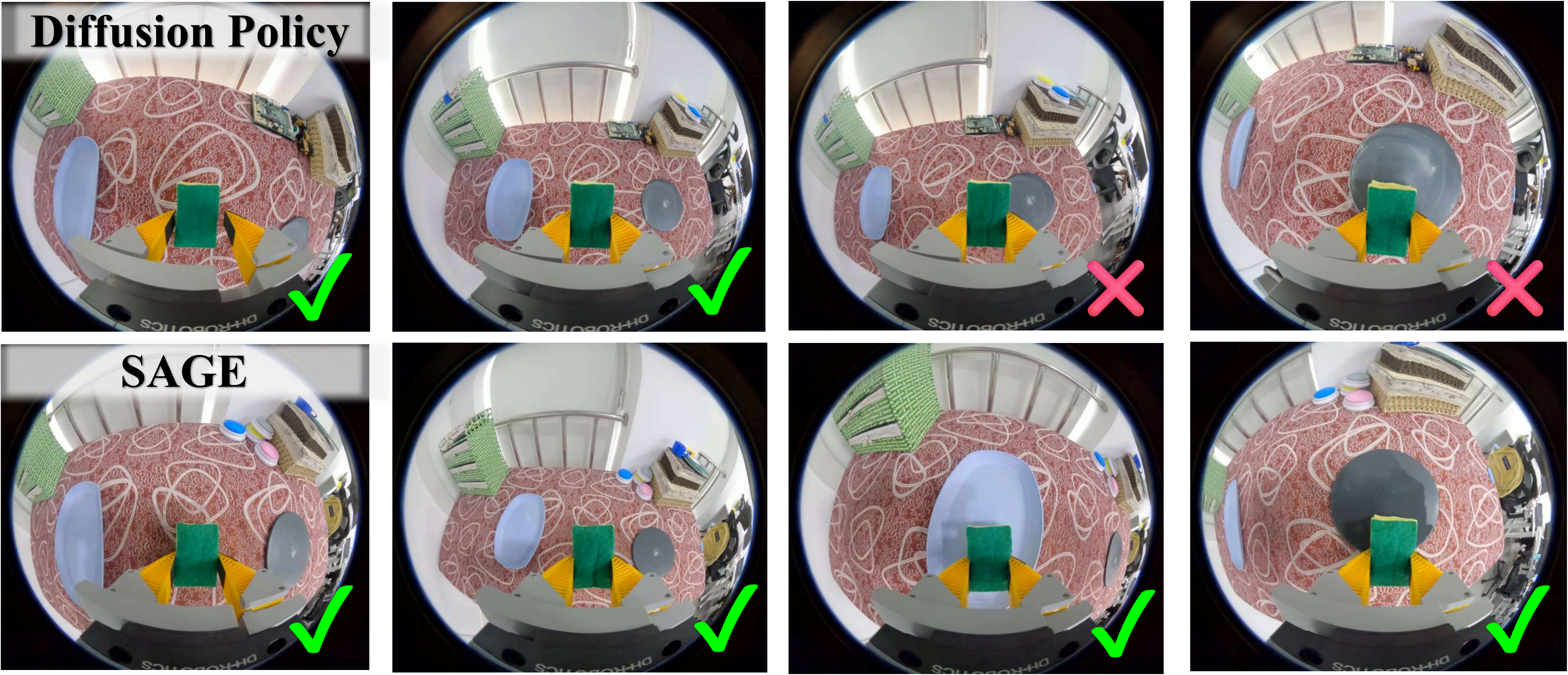}
    \caption{Visualizations of success and failure cases in the \textit{Wipe Dish} task.}
    \label{fig:Wash-dish-demo}  
    \vspace{-0.5cm}
\end{figure}

\subsection{Stage-Level Analysis of Each Task}

\textbf{Evaluation Setting.}
In this section, we conduct stage-level evaluations based on the tasks illustrated in Fig.~\ref{fig:Task-Design}. Each configuration is tested independently over 20 trials. For each trial, we record the cumulative success rate at every stage. Specifically, the success rate at stage $i$ is defined as the proportion of trials in which all stages from the initial state up to stage $i$ are completed successfully in the correct order. Full results are reported in Table~\ref{tab:all_tasks_compact}.

The \textit{Sort Blocks} and \textit{Push Buttons} tasks are further extended into distracting conditions: \textit{Sort Blocks} introduces additional yellow and green blocks. \textit{Push Buttons} introduces an additional gray button. These new colors do not appear in the training set, serving as out-of-distribution distractors.

\begin{table}[t]
\centering
\renewcommand{\arraystretch}{0.8}
\setlength{\tabcolsep}{5pt}
\caption{Per-Stage Success Rate Across Tasks}
\label{tab:all_tasks_compact}
\begin{tabular}{p{1.7cm}|c|cccc}
\toprule
\textbf{Task} & \textbf{Model} & \textbf{Stage 1} & \textbf{Stage 2} & \textbf{Stage 3} & \textbf{Stage 4} \\
\midrule
\multirow{4}{=}{Sort Blocks\\(standard)} 
    & DP           & 0.65  & 0.50  & 0.30  & 0.25 \\
    & BC-RNN       & 0.05  & 0.00   & 0.00   & 0.00  \\
    & Hierarchical  & 0.95  & 0.85  & 0.80  & 0.65 \\
    & \name{}         & \textbf{1.00} & \textbf{1.00} & \textbf{1.00} & \textbf{1.00} \\
\midrule
\multirow{4}{=}{Sort Blocks\\(distracting)} 
    & DP           & 0.55  & 0.40  & 0.20  & 0.15 \\
    & BC-RNN       & 0.00   & 0.00   & 0.00   & 0.00  \\
    & Hierarchical  & 0.85  & 0.85  & 0.80  & 0.55 \\
    & \name{}         & \textbf{0.95}  & \textbf{0.90} & \textbf{0.85} & \textbf{0.80} \\
\midrule
\multirow{4}{=}{Push Buttons\\(standard)} 
    & DP           & 0.80  & 0.60  & 0.40  & --   \\
    & BC-RNN       & 0.10   & 0.05   & 0.00   & --   \\
    & Hierarchical  & \textbf{1.00} & 0.70  & 0.60  & --   \\
    & \name{}         & \textbf{1.00} & \textbf{1.00} & \textbf{1.00} & -- \\
\midrule
\multirow{4}{=}{Push Buttons\\(distracting)} 
    & DP           & 0.80  & 0.40  & 0.30  & --   \\
    & BC-RNN       & 0.00   & 0.00   & 0.00   & --   \\
    & Hierarchical  & \textbf{1.00} & 0.65  & 0.55  & --   \\
    & \name{}         & \textbf{1.00} & \textbf{1.00} & \textbf{1.00} & -- \\
\midrule
\multirow{4}{=}{Wipe Dish\\(standard)} 
    & DP            & \textbf{1.00} & 0.40 & 0.35  & -- \\
    & BC-RNN        & 0.10   & 0.00  & 0.00   & -- \\
    & Hierarchical   & 0.85  & 0.85 & 0.75  & -- \\
    & \name{}          & \textbf{1.00} & \textbf{1.00} & \textbf{1.00} & -- \\
\bottomrule

\end{tabular}
\end{table}

\textbf{Result Analysis.}
Experimental results show that our method consistently outperforms all baselines across tasks and conditions. This confirms the effectiveness of our HMDP model in disambiguating visually similar observations across stages and maintaining high decision accuracy throughout long-horizon execution.

The \textbf{DP} model, which relies exclusively on short-term visual inputs, exhibits a strong spatial bias: It tends to select objects that appear closer in the egocentric view, regardless of task-specific semantics. For instance, in the \textit{Sort Blocks} task, DP often grasps the nearest block—even if it is a distractor such as yellow or green, which never appeared during training—thereby ignoring the required red–blue sequence. In the \textit{Wipe Dish} task, DP performs perfectly in Stage 1 (success rate: 1.00), where visual cues are unambiguous. However, its success rate drops sharply to 0.40 at Stage 2, where state ambiguity arises—the model must decide whether to dip the cloth into water or wipe the dish, but dry and wet cloths appear visually identical. Without stage-awareness, DP tends to make near-random decisions between dipping and wiping; for instance, if the cloth appears closer to the dish, it may skip the soaking step and proceed directly to wiping (as shown in Fig. \ref{fig:Wash-dish-demo}), ultimately leading to task failure. This highlights its inability to resolve state ambiguity from visual input alone.

The \textbf{Hierarchical} approach lags behind \name{}, likely due to its rigid assumptions about stage boundaries—requiring each action to be clearly categorized as a primitive~\cite{luo2024multistage} or skill module~\cite{silver2023learning}—which may be unrealistic in some real-world tasks. For example, in the \textit{Sort Blocks} task, completing a single stage involves multiple low-level steps—approaching the block, grasping, lifting, locating the tray, and releasing—all of which may exceed the scope of a typical primitive. As a result, the model frequently switches stages prematurely or fails to detect necessary transitions, leading to execution errors and incomplete behaviors.

The \textbf{BC-RNN} model attempts to capture long-term dependencies through recurrence. However, it suffers from noisy visual input and lacks explicit stage-awareness. In the \textit{Push Buttons} task, for example, the model must infer which button to press next based on visual history, but the input sequence often contains redundant or visually similar frames, making it difficult to extract relevant temporal cues. As a result, BC-RNN frequently exhibits prolonged inaction or oscillatory behavior, unable to determine whether the current button has already been pressed—ultimately leading to poor stage completion rates.

In contrast, our method achieves consistently high success rates across all stages and tasks, including distracting conditions. This demonstrates its robustness to \textit{state ambiguity} and its ability to model task progression without relying on task-specific high-level controllers.

\subsection{Infinite-Horizon: Stress Testing Sequential Robustness}

To evaluate the long-horizon stability and robustness of methods, we introduce an extended version of the \textit{Sort Blocks} task. In this setting, the workspace is continuously replenished with blocks by a human operator standing nearby, such that there are always at least two red and two blue blocks available. The robot is required to repeatedly execute the target sequence: \textbf{red $\rightarrow$ blue $\rightarrow$ red $\rightarrow$ blue $\rightarrow$ ...}, without reset between episodes. 

Each trial continues until one of the following conditions is met:
(1) The robot grasps a block in the wrong order (violating the red–blue alternation), or (2) the robot successfully completes 50 consecutive grasps following the correct sequence.

This setting serves as a stress test for the model’s stage tracking ability under indefinite execution. We evaluate 3 methods with 3 independent trials and report the average number of successful consecutive grasps before failure (up to a maximum of 50). The results are summarized in Table~\ref{tab:long_horizon_stability}.

\begin{table}[t]
\centering
\caption{Average number of correct consecutive grasps in extended Sort Blocks task (max 50).}
\label{tab:long_horizon_stability}
\begin{tabular}{l|cccc}
\toprule
\textbf{Model} & \textbf{Trial 1} & \textbf{Trial 2} & \textbf{Trial 3} & \textbf{Average} \\
\midrule
DP & 4 & 1 & 3 & 2.7 \\
Hierarchical & 3 & 8 & 4 & 5.0 \\
\name{} & \textbf{49} & \textbf{50} & \textbf{50} & \textbf{49.7} \\
\bottomrule
\end{tabular}
\vspace{-0.3cm}
\end{table}

Our method achieves near-perfect performance across all three trials, consistently completing 50-step sequences without error in two of them. This confirms its strong capability to maintain task progression and resolve \textit{state ambiguity} over prolonged horizons. In contrast, other baselines either violate the sequence prematurely or exhibit inconsistent performance under repeated cycles.

\section{Ablation Study}
\subsection{Effect of Sampling Strategy and Label Type}
\label{subsec:ablation-soft}

In Section~\ref{sec:auto_tag}, we propose a semi-automatic labeling approach for state annotation, aiming to reduce manual effort. To validate its effectiveness, we conduct ablation studies on two key aspects: (1) Sample selection (active learning vs. random sampling), and (2) label formulation (soft vs. hard). The annotation strategy ultimately provides state supervision for our HMDP-based state-aware action policy, so the study examines its impact on both annotation quality and downstream task performance.

\textbf{Evaluation Process.} We evaluate our method on the \textit{Sort Blocks} task, which contains a total of 157 episodes. All episodes are manually labeled once to serve as ground-truth annotations for evaluation. For training, however, we randomly select $b$ episodes for manual labeling, train a state auto-label model on this subset, and use it to annotate the remaining episodes. The resulting combination of manually labeled ($b$) and automatically generated ($157\!-\!b$) labels is then used to train the End-to-End model following the procedure described in Section~\ref{subsec:end-to-end}.

\begin{figure}[!t]
    \centering
    \includegraphics[width=1\linewidth]{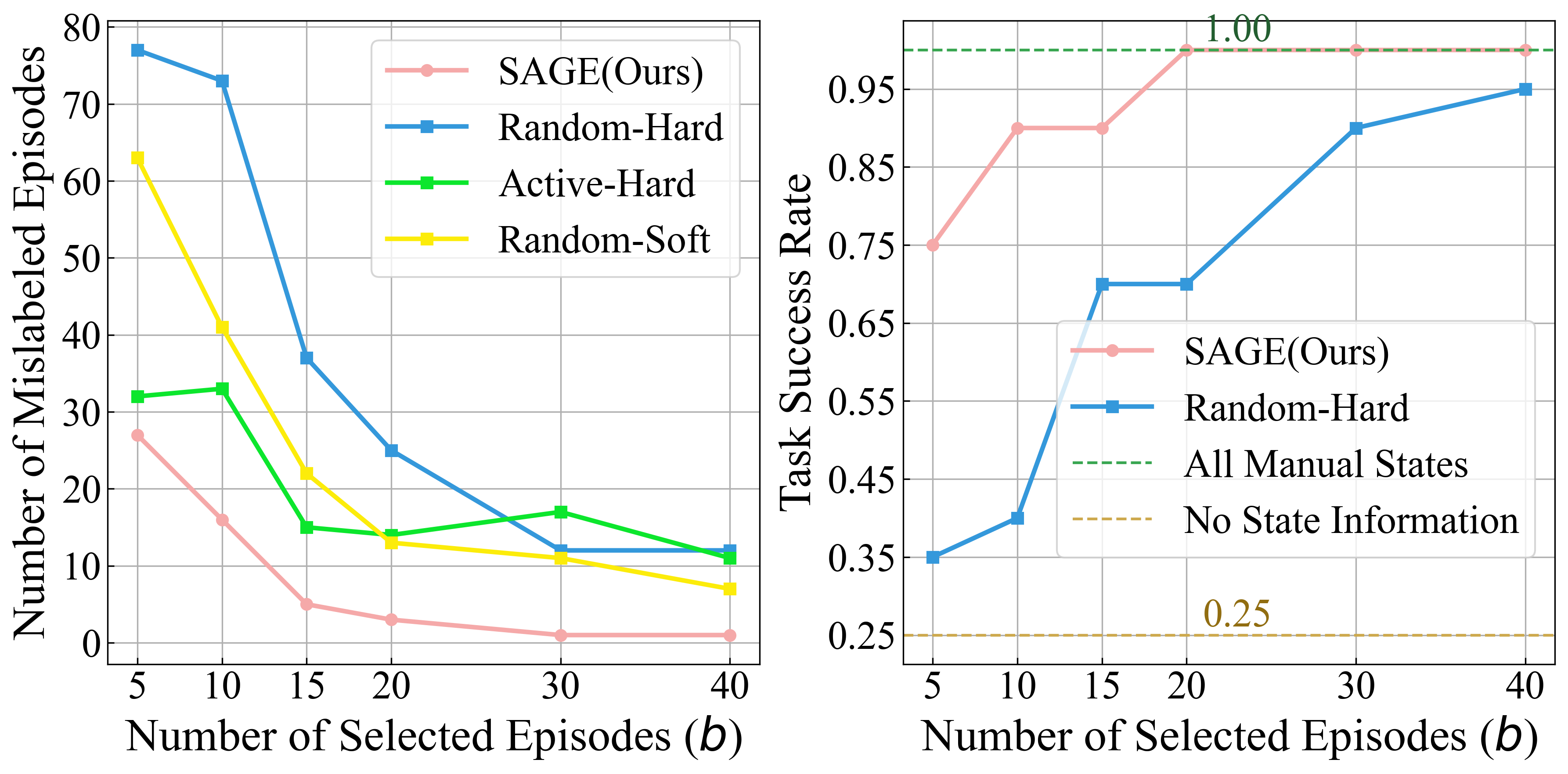}
    \vspace{-0.5cm}
    \caption{Ablation study on annotation strategies using the \textit{Sort Blocks} task. \textbf{Left:} Number of mislabeled episodes (accuracy $<$ 90\%) versus number of selected samples $b$. \textbf{Right:} Task success rate of policies trained on the resulting auto-labeled data. Dashed lines indicate upper (fully manual) and lower (no state) supervision baselines.}
    \label{fig:ablation}
    \vspace{-0.5cm}
\end{figure}

\textbf{Evaluation Metrics.} We assess both annotation quality and downstream task performance. For annotation quality, predictions are compared against human ground-truth labels: A test episode is considered \textit{mislabeled} if its state prediction accuracy falls below 90\%. We report the number of mislabeled episodes for each method (Fig.~\ref{fig:ablation}, left). For task performance, we train the policy with the mixed labels and measure the success rate on the \textit{Sort Blocks} task (Fig.~\ref{fig:ablation}, right). We evaluate four variants covering all combinations of \textit{sample selection} (random vs. active) and \textit{label formulation} (hard vs. soft).

$\bullet$ \emph{SAGE (Ours).} Selects episodes actively using active learning (Section~\ref{subsec:active_tag}), interpolates human annotations into soft state labels (Eq.~\eqref{eq:soft-label}), and trains the annotation model with KL-divergence reward (Eq.~\eqref{eq:loss-AL-kl}).

$\bullet$ \emph{Active-Hard.} Selects episodes actively using active learning (Section~\ref{subsec:active_tag}), annotates them with hard (discrete) labels, and trains the annotation model with standard cross-entropy reward(Eq.~\eqref{eq:L1}).

$\bullet$ \emph{Random-Soft.} Selects episodes randomly, interpolates human annotations into soft state labels, and trains the annotation model with KL-divergence reward (Eq.~\eqref{eq:loss-AL-kl}).

$\bullet$ \emph{Random-Hard.} Selects episodes randomly, annotates them with hard (discrete) labels, and trains the annotation model with standard cross-entropy reward (Eq.~\eqref{eq:L1}).

\textbf{Result Analysis.} 
Fig.~\ref{fig:ablation} presents the performance of all methods across different values of $b$. SAGE consistently yields the fewest mislabeled episodes and the highest task success rate. Remarkably, with only about 10 annotated episodes, SAGE achieves 90\% task success, approaching the performance of full manual labeling. With 20 annotated episodes, SAGE reaches 100\% success, matching full supervision while requiring only around 13\% of the total episodes to be manually labeled, demonstrating both annotation efficiency and high-quality guidance.

The effects of the two key factors—\textit{sample selection} and \textit{label formulation}—can be seen by comparing across the baselines. Comparing SAGE with Random-Soft (both using soft labels) highlights the benefit of active learning: Representative episode selection improves generalization beyond what soft labels alone can achieve, especially at low $b$. Comparing SAGE with Active-Hard (both using active learning) isolates the effect of soft labels: Interpolated soft labels stabilize training and improve performance. Random-Hard, which lacks both active learning and soft supervision, performs worst, confirming that the two factors provide complementary improvements.

We also observe a saturation effect: Once the number of mislabeled episodes drops below roughly 10, additional annotations yield diminishing returns in task success. This suggests that the policy can tolerate minor annotation noise, likely due to short transition windows.

In summary, these results demonstrate that SAGE’s combination of active learning and soft labeling achieves superior annotation efficiency and robust downstream policy performance, requiring only around 13\% of the manual labeling effort compared to fully supervised approaches.

\subsection{Sensitivity Analysis on Soft Label Transition Length}

\begin{figure}[!t]
    \centering
    \includegraphics[width=0.35\textwidth]{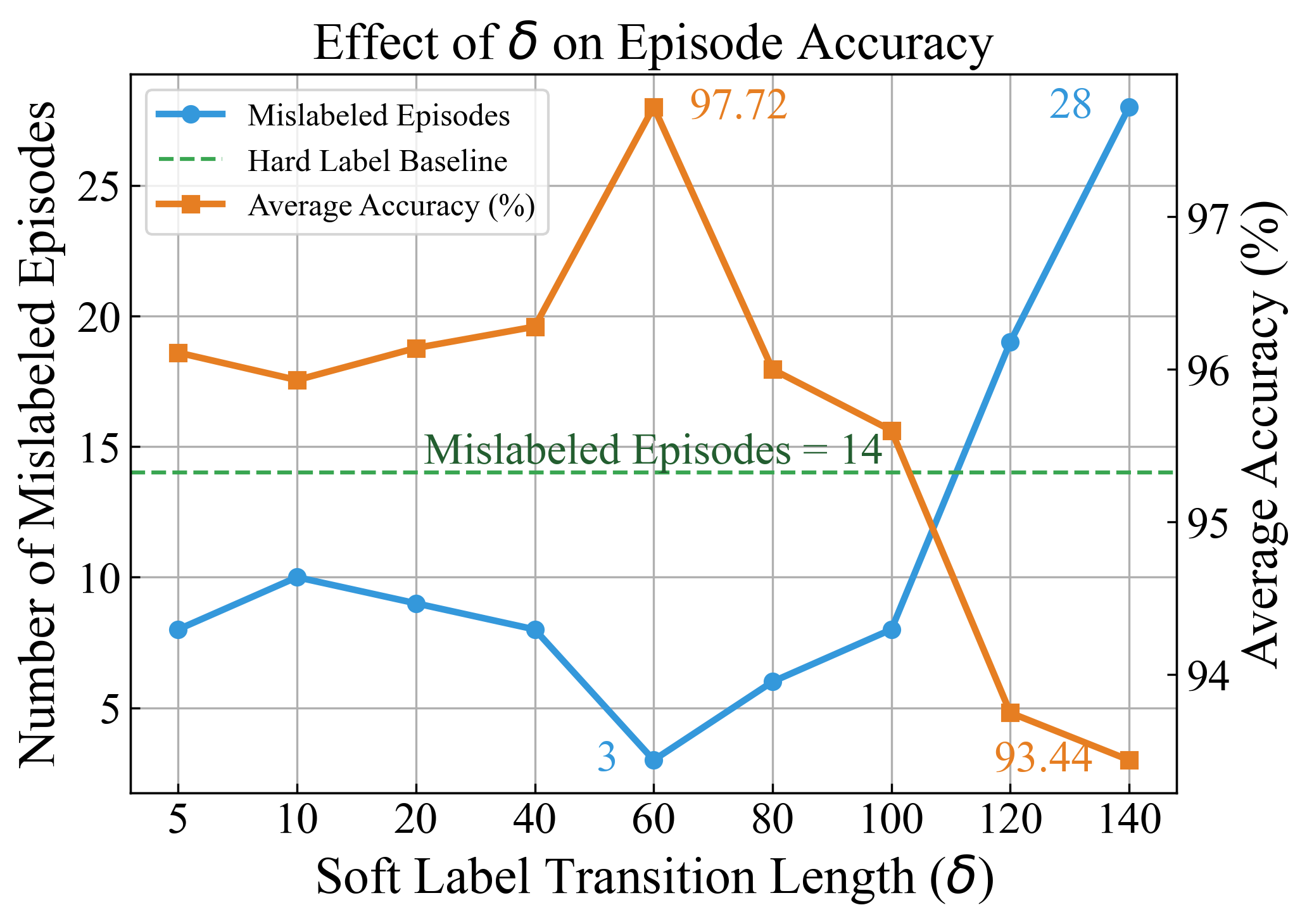}
    \caption{Effect of soft label transition length $\delta$ on annotation quality and model performance. Left y-axis: Number of mislabeled episodes (accuracy $<$ 90\%). Right y-axis: Average episode accuracy (\%). Green dashed line indicates the baseline performance of hard labels ($\delta=0$).}
    \label{fig:delta_ablation}
    \vspace{-0.5cm}
\end{figure}

To isolate the effect of soft label transition length $\delta$, we perform a controlled ablation on the same 20 training episodes, all selected via active learning(Section ~\ref{subsec:active_tag}). All other conditions (model architecture, optimization steps, and training epochs) are held constant. The only variable is $\delta$, controlling the width of the temporal transition zone for soft state label interpolation (Eq.~\eqref{eq:soft-label}).

We vary $\delta$ from 5 to 140 frames and report two metrics in Fig.~\ref{fig:delta_ablation}: Mislabeled episodes (accuracy $<$ 90\%) and average validation accuracy. The green dashed line indicates hard-label training ($\delta=0$).

We observe a U-shaped trend in misannotation rates: Performance improves with increasing $\delta$, peaks at $\delta=60$ (about 1 second), and degrades for larger values. Average accuracy follows a similar pattern, reaching 97.72\% at $\delta=60$. Most soft label settings ($\delta \in [5,100]$) perform comparably or better than hard labels, confirming the robustness of soft state interpolation.

The benefit of soft labels comes from expanding the state transition into a temporal corridor, providing smoother supervision and enabling the model to learn gradual state changes. Too large $\delta$ causes over-smoothing, while too small $\delta$ loses the advantages of smooth guidance.

In summary, a moderate transition length ($\delta=60$) achieves the best trade-off, validating the effectiveness and tolerance of soft label smoothing.

\subsection{State Annotation Strategies}
\label{subsec:state-annotation}

\begin{figure*}[t]
    \centering
    \begin{minipage}[t]{0.48\textwidth}
        \centering
        \includegraphics[width=\linewidth]{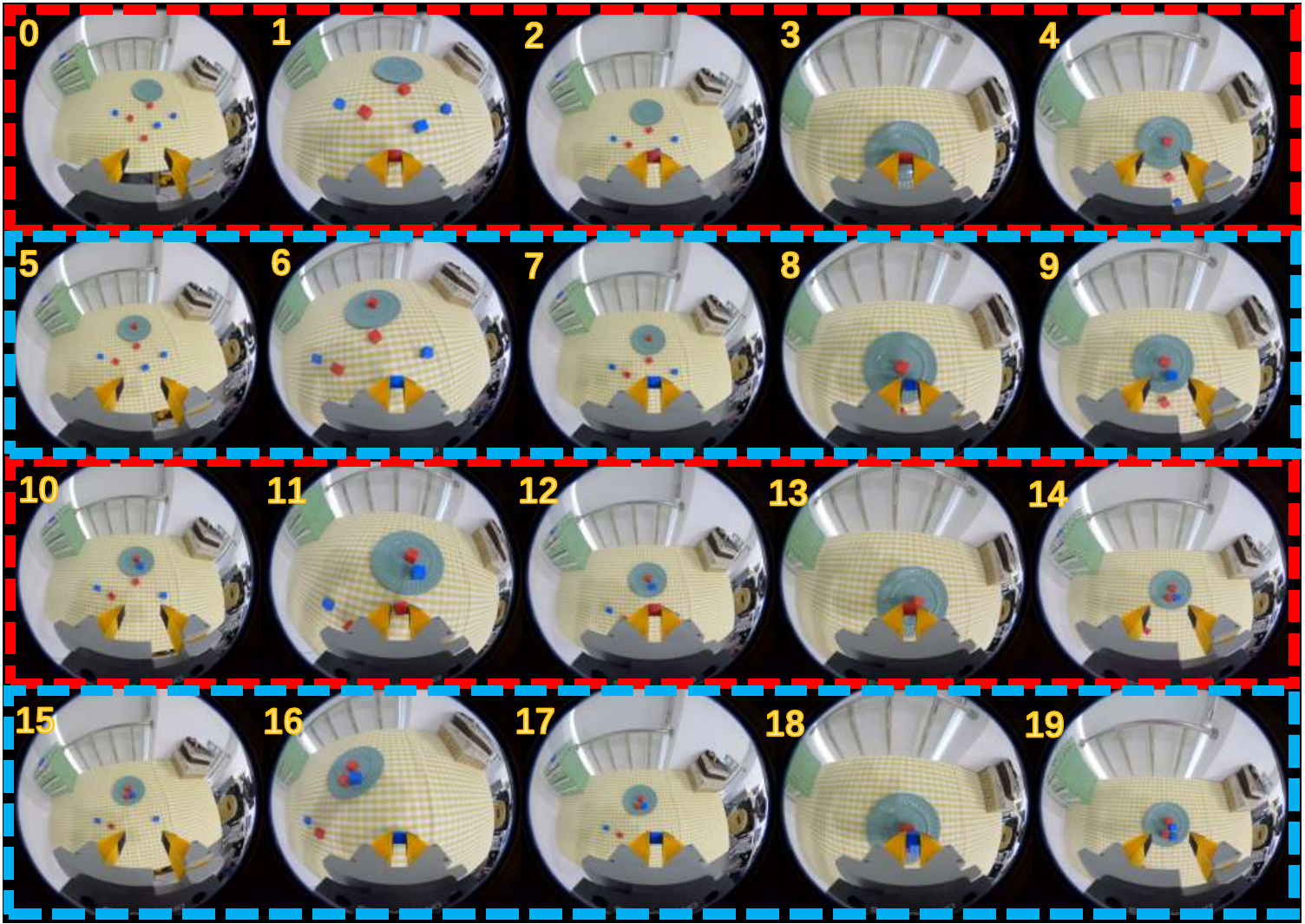}
        \caption*{(a)Post-Placement Boundary}
    \end{minipage}
    \hfill
    \begin{minipage}[t]{0.48\textwidth}
        \centering
        \includegraphics[width=\linewidth]{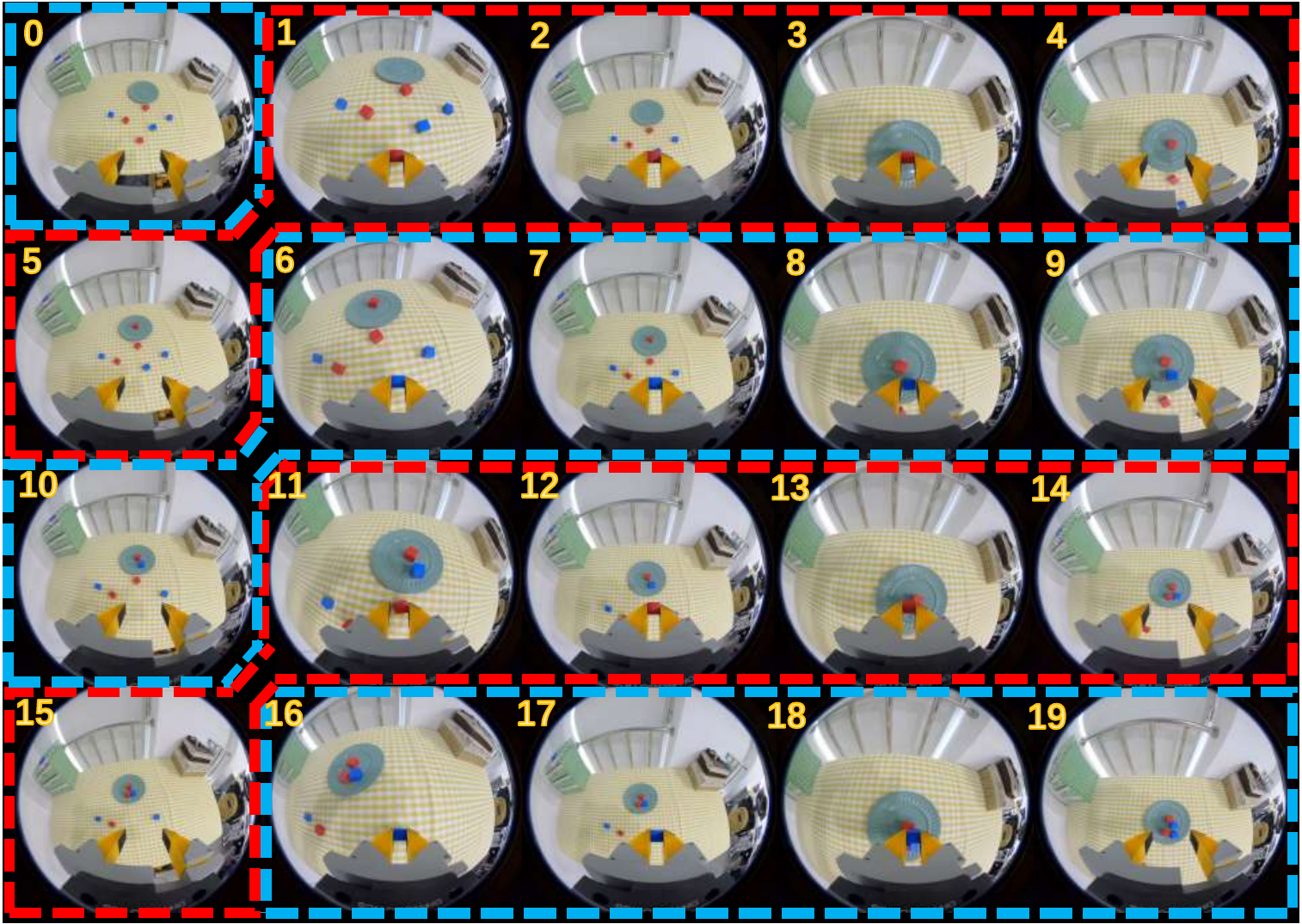}
        \caption*{(b)Pre-Grasp Boundary}  
    \end{minipage}
    \caption{
        Comparison of two state boundary definitions. Time steps within the same dashed box are annotated as belonging to the same hidden state.
        \textbf{Left:} Post-Placement Boundary—each state starts after placing the previous block and includes the next grasp.
        \textbf{Right:} Pre-Grasp Boundary—each state begins at the start of a new grasp, excluding the previous placement.
    }
    \label{fig:boundary-comparison}
    \vspace{-0.3cm}
\end{figure*}

\begin{table}[t]
\centering
\caption{Success rate of SAGE under different annotation strategies.}
\label{tab:annotation-ablation}
\begin{tabular}{l|c}
\toprule
\textbf{Annotation Strategy} & \textbf{Success Rate} \\
\midrule
Post-Placement Boundary & 1.00 (20/20) \\
Pre-Grasp Boundary      & 1.00 (20/20) \\
\bottomrule
\end{tabular}
\vspace{-0.3cm}
\end{table}

In this experiment, we examine whether the specific definition of stage boundaries affects the performance of our method. Since our approach models task stages via inferred hidden states, it is natural to question whether the precise rule used to define such states during training impacts the learned policy, especially under state ambiguity.

To investigate this, we conduct an ablation study on the \textit{Sort Blocks} task, comparing two distinct annotation strategies for stage segmentation. As illustrated in Fig.~\ref{fig:boundary-comparison}, the two strategies differ in the temporal location of the boundary transition:

$\bullet$ \emph{Post-Placement Boundary:} The state is switched immediately after placing a block. Each stage spans from the end of the previous placement, through the grasp and lift, until the next placement finishes, as shown in Fig.~\ref{fig:boundary-comparison} (a).
    
$\bullet$ \emph{Pre-Grasp Boundary:} The state is switched at the onset of grasping a new block. Each stage begins when the robot reaches for the next block and ends upon its successful grasp, as shown in Fig.~\ref{fig:boundary-comparison} (b).

While both strategies cover the full manipulation sequence, they differ in how transitional phases are grouped. Crucially, both apply a consistent and temporally structured rule for state assignment. We train our proposed state-aware action policy separately using each annotation scheme and evaluate the learned policies over 20 test episodes. As shown in Table~\ref{tab:annotation-ablation}, both models achieve a 100\% success rate, demonstrating strong robustness to annotation strategies.

These findings indicate that, unlike hierarchical approaches that rely on task-specific stage boundary rules~\cite{luo2024multistage,HVIL,gupta2019relay}, the effectiveness of our framework does not depend on such heuristics. Instead, it leverages the hidden state inference module to adapt to the underlying task dynamics.

\subsection{Evaluating the Action Dependency on Hidden States}

\begin{table}[t]
\centering
\caption{Injected state sequences and corresponding execution results.}
\label{tab:state_action_dependency}
\begin{tabular}{c|p{5.5cm}|c}
\toprule
\textbf{Trial ID} & \textbf{Injected State Sequence ($s_t^{\text{fix}}$)} & \textbf{Result} \\
\midrule
1 & \texttt{0 0 1 1 0 1 0 1 1 0 0 0 0 0 1} & All Correct \\
2 & \texttt{1 0 1 1 0 0 1 0 1 0 1 1 0 0 1} & All Correct \\
3 & \texttt{0 1 0 1 1 0 0 1 1 0 0 1 0 1 0} & All Correct \\
\bottomrule
\end{tabular}
\vspace{-0.3cm}
\end{table}

To directly assess the influence of hidden state representations on the model's action decisions, we design an intervention-based ablation in the \textit{Sort Blocks} task. This experiment aims to answer a fundamental question: \textit{Does the state-aware action policy truly condition its actions on the inferred state}?

We construct a controlled environment where both red and blue blocks are always available on the table. The robot no longer follows a pre-defined red–blue sequence. During inference, we manually override the model’s inferred state by injecting a fixed scalar state $s_t^{\text{fix}} \in \{0,1\}$ into the policy, replacing $\hat{\bm{s}}_t$ in Eq.~\eqref{eq:a=pi}. This scalar is internally mapped to a one-hot vector: $s_t^{\text{fix}}=0$ corresponds to $(1,0)^T$ (“red block stage”), and $s_t^{\text{fix}}=1$ corresponds to $(0,1)^T$ (“blue block stage”). Consequently, the policy receives identical visual observations across trials, while the state input differs according to the injected scalar $s_t^{\text{fix}}$.

Each trial consists of 15 steps, following a random state sequence (e.g., 001101 $\rightarrow$ red, red, blue, blue, red, blue). The goal is to evaluate whether the policy obeys the externally injected state and executes the correct color-conditioned grasp.

We perform three such trials and summarize the results in Table~\ref{tab:state_action_dependency}, where each trial injects a random 15-step state sequence. The robot is expected to grasp a red block for state 0 and a blue block for state 1.

These results provide strong evidence that the learned policy meaningfully integrates state information when generating actions. Despite having access to the same egocentric visual input in all cases, the model consistently adapts its behavior in accordance with the injected state. This confirms that the hidden state variable $\bm{s}_t$ plays a causal and decisive role in guiding the policy’s action generation, rather than acting as a redundant feature.

\section{Conclusion and Future Work}
This paper proposes SAGE, an End-to-End imitation learning framework that incorporates hidden state modeling to guide long-horizon decision-making. Based on the HMDP formulation, SAGE infers hidden states from visual observations and previous hidden states. It then uses these hidden states to resolve state ambiguity and ensure consistent policy execution. To reduce the burden of supervision for states, we introduce a semi-automatic labeling pipeline that combines active learning-based episode sampling and soft label refinement, requiring only around 13\% of the total episodes to be manually labeled. Comprehensive experiments demonstrate that SAGE consistently outperforms baselines in both accuracy and robustness, confirming the advantage of embedding structured hidden state representations into the policy learning process. 

Future work can aim to further reduce or completely eliminate human supervision. Although the current approach already significantly saves labeling effort through active learning-based episode sampling and soft label refinement, a small number of episodes still require manual annotation to train the state auto-label model. To achieve fully automatic stage segmentation and state labeling, future research could leverage additional information from video sequences and recorded actions, enabling the model to infer task stages and hidden states directly from raw demonstrations without any human-provided labels. This would further enhance the scalability and applicability of SAGE to large-scale or continuously evolving tasks.


\begingroup
\balance
\bibliographystyle{refactored-bst}

\endgroup

\newpage

\end{document}